\documentclass[runningheads]{llncs}
\usepackage[utf8]{inputenc}
\usepackage[noblocks]{authblk}
\usepackage{geometry}
\usepackage{amsmath}

\usepackage{amsthm,cite}
\usepackage[caption=false]{subfig}
\usepackage{amssymb}
\usepackage{graphicx}
\usepackage{picture}
\usepackage{geometry}
\usepackage{enumerate}
\usepackage{algorithm}
\usepackage{algorithmic}
\usepackage{dsfont}
\usepackage{bm,xr}
\usepackage{tikz}
\usepackage{collcell,url}
\usepackage{colortbl}
\usepackage{rotating,booktabs}
\def\ddfrac#1#2{\displaystyle\frac{\displaystyle #1}{\displaystyle #2}}

\DeclareMathOperator*{\argmin}{arg\,min}
\usepackage{sectsty}
\sectionfont{\fontsize{11}{12}\selectfont}
\subsectionfont{\fontsize{9}{10}\selectfont}
\subsubsectionfont{\fontsize{9}{10}\selectfont}

\let\OLDthebibliography\thebibliography
\renewcommand\thebibliography[1]{
  \OLDthebibliography{#1}
  \setlength{\parskip}{0pt}
  \setlength{\itemsep}{0pt plus 0.3ex}
}

\allowdisplaybreaks[3]
 
\newcommand{\cP}{\mathcal{P}}

\begin{document}
\title{\textbf{A Ranking Model Motivated  by Nonnegative Matrix Factorization with Applications to Tennis Tournaments}}

\author{Rui Xia\inst{1} \and
Vincent Y. F. Tan\inst{1,2} \and
Louis~Filstroff\inst{3} \and
C\'edric F\'evotte\inst{3}}

\authorrunning{R.~Xia, V.~Y.~F.~Tan, L.~Filstroff and C.~F\'evotte}
\titlerunning{ }
\institute{Department of Mathematics, National University of Singapore (NUS)  \and Department of Electrical and Computer Engineering, NUS\\ \email{rui.xia@u.nus.edu, vtan@nus.edu.sg} \and IRIT, Universit\'e de Toulouse, CNRS, France \\
\email{\{louis.filstroff, cedric.fevotte\}@irit.fr}}

\maketitle
\begin{abstract}
We propose a novel    ranking model that combines the Bradley-Terry-Luce  probability model with a nonnegative matrix factorization  framework to model and uncover the presence of latent variables  that influence the performance of top tennis players. We derive an efficient, provably convergent, and numerically stable  majorization-minimization-based algorithm  to maximize the likelihood of datasets under the proposed statistical model. The model is tested on datasets involving the outcomes of matches between $20$ top male and female tennis players over $14$   major tournaments for men (including   the Grand Slams and   the  ATP Masters 1000)  and    $16$ major tournaments  for women over the past $10$ years. Our model automatically infers that the  surface  of the court (e.g., clay or hard court) is a key determinant of the performances of male players, but less so for females. Top players on various surfaces over this longitudinal period are also identified in an objective manner. \\

\end{abstract}

\section{Introduction}
The international rankings for both male and female tennis players  are based on a rolling $52$-week, cumulative system, where ranking points are earned from players' performances at tournaments. However, due to the limited observation window, such a ranking system is not sufficient if one would like to compare  dominant players over a long period (say $10$ years)  as   players peak at different times.  The ranking points that players accrue depend only on the stage of the tournaments reached by him or her. Unlike the well-studied Elo rating system for chess~\cite{EloBook}, one opponent's ranking is not taken into account, i.e., one will not be awarded with  bonus points by defeating a top player.  Furthermore, the current ranking system does not take into account the players' performances under different conditions (e.g., surface type of courts). We propose a statistical  model to ameliorate the above-mentioned shortcomings by  (i) understanding the relative ranking of players over a longitudinal period and (ii) discovering the existence of any latent variables that influence players' performances. 

The statistical model we propose is an amalgamation of two well-studied models in the ranking and dictionary learning literatures, namely, the {\em Bradley-Terry-Luce} (BTL) model~\cite{bradleyterry, Luce} for ranking a population of items (in this case, tennis  players)  based on pairwise comparisons and {\em nonnegative matrix factorization} (NMF)~\cite{LS99,CichockiBook}. The BTL model posits that given  a pair of players $(i,j)$ from a population of players $\{1,\ldots, N\}$, the probability that the pairwise comparison ``$i$ beats $j$'' is true is given by
\begin{equation}
    \Pr(i  \;\text{beats}\; j) = \frac{\lambda_{i}}{\lambda_{i}+\lambda_{j}} .\label{eqn:btl}
\end{equation}
Thus, $\lambda_i\in\mathbb{R}_+ :=[0,\infty)$  can be interpreted as the {\em skill level} of player $i$.  The row vector ${\bm \lambda}=(\lambda_1,\ldots, \lambda_N)\in\mathbb{R}_+^{1\times N}$ thus parametrizes the BTL model.  Other more general ranking models are discussed in \cite{MardenBook} but the BTL model suffices as the outcomes of tennis matches are binary.

NMF   consists in the following problem. Given a nonnegative matrix $\mathbf{\Lambda} \in\mathbb{R}_+^{M\times N}$, one would like to find two matrices $\mathbf{W}\in\mathbb{R}_+^{M\times K}$ and $\mathbf{H}\in\mathbb{R}_+^{K\times N}$ such that  their product $\mathbf{W}\mathbf{H}$ serves as a good low-rank approximation to $\mathbf{\Lambda}$. NMF is a linear dimensionality reduction technique that has seen a surge in popularity since the seminal papers by Lee and Seung~\cite{LS99,leeseung2000}.  Due to the non-subtractive nature of the decomposition,  constituent parts of objects can be extracted from complicated datasets. The matrix $\mathbf{W}$, known as the {\em dictionary matrix}, contains in its columns the parts, and the matrix $\mathbf{H}$, known as the {\em coefficient matrix}, contains in its rows  activation coefficients that encode how much of each part is present in the columns of the data matrix $ \mathbf{\Lambda}$. NMF has also been used successfully to uncover latent variables with specific interpretations in various applications, including audio signal processing~\cite{fevotte2009}, text mining analysis \cite{berry2005}, and even analyzing soccer players' playing style~\cite{geerts2018}. We combine this framework with the BTL model to perform a   {\em sports analytics} task on top tennis players. % in this paper. 

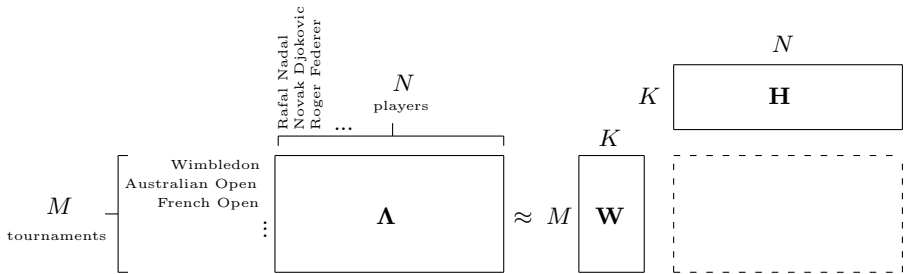
\begin{figure}[t]
\centering
\setlength{\unitlength}{.43mm}
\begin{picture}(300, 80)

\put(0,10){\tiny tournaments}
\put(12,18){\small $M$}
\put(30,18){\line(1,0){4}}
\put(34,0){\line(0,1){36}}
\put(34,36){\line(1,0){3}}
\put(34,0){\line(1,0){3}}
\put(50,32){\tiny Wimbledon}
\put(36,26){\tiny Australian Open}
\put(46,20){\tiny French Open}
\put(78,15){.}
\put(78,13){.}
\put(78,11){.}

\put(82,0){\line(0,1){36}}
\put(82,36){\line(1,0){70}}
\put(82,0){\line(1,0){70}}
\put(152,0){\line(0,1){36}}
\put(113,15){\small $\mathbf{\Lambda}$}

\put(83,42){\line(1,0){69}}
\put(83,42){\line(0,-1){3}}
\put(152,42){\line(0,-1){3}}
\put(118,42){\line(0,1){4}}
\put(118,56){\small $N$}
\put(112,50){\tiny players}
\put(83,44){\rotatebox{90}{\tiny Rafal Nadal}}
\put(88,44){\rotatebox{90}{\tiny Novak Djokovic}}
\put(93,44){\rotatebox{90}{\tiny Roger Federer}}
\put(100,44){.}
\put(102,44){.}
\put(104,44){.}

\put(155, 15){\small $\approx$}

\put(165, 15){\small $M$}
\put(175, 0){\line(0,1){36}}
\put(175, 0){\line(1,0){20}}
\put(175,36){\line(1,0){20}}
\put(195, 0){\line(0,1){36}}
\put(181,39){\small $K$}
\put(180,15){\small $\mathbf{W}$}

\multiput(204,0)(5,0){14}{\line(1,0){2}}
\multiput(204,0)(0,4.5){8}{\line(0,1){2}}
\multiput(204,36)(5,0){14}{\line(1,0){2}}
\multiput(274,0)(0,4.5){8}{\line(0,1){2}}

\put(234, 68){\small $N$}
\put(204, 44){\line(1,0){70}}
\put(204, 44){\line(0,1){20}}
\put(204, 64){\line(1,0){70}}
\put(274, 44){\line(0,1){20}}
\put(193, 52){\small $K$}
\put(233, 52){\small $\mathbf{H}$}
\end{picture}
	\caption{The  BTL-NMF Model}
	\label{fig:nmf_btl}
\end{figure}

\subsection{Main Contributions}
\paragraph{Model:} In this paper, we amalgamate the aforementioned models to rank tennis players and uncover latent factors that influence their performances. We propose a hybrid {\em  BTL-NMF} model (see Fig.~\ref{fig:nmf_btl}) in which there are $M$ different  skill vectors $\bm{\lambda}_m, m \in \{1,\ldots, M\}$, each representing players' relative skill levels in various tournaments indexed by $m$. These row vectors are stacked into an $M\times N$ matrix $\mathbf{\Lambda}$ which is the given input matrix in an NMF model. 

\paragraph{Algorithms and Theory:} We develop computationally efficient and numerically stable majorization-minimization (MM)-based algorithms \cite{hunterLange2004}  to obtain a  decomposition of  $\mathbf{\Lambda}$  into  $\mathbf{W}$ and $\mathbf{H}$ that maximizes the likelihood of the data. Furthermore, by using ideas from~\cite{zhao2017unified,Raz}, we prove that not only is the objective function monotonically non-decreasing along iterations, additionally, every limit point of the sequence of iterates of the dictionary and coefficient matrices is  a {\em stationary point} of the objective function.

\paragraph{Experiments:} We collected rich datasets of pairwise outcomes of $N=20$ top male and female players and $M=14$ (or $M=16$) top tournaments over $10$ years. Based on these datasets, our algorithm yielded factor matrices $\mathbf{W}$ and $\mathbf{H}$ that allowed us to draw interesting conclusions about the existence of latent variable(s) and relative rankings of dominant   players over the past $10$ years. In particular, we conclude that male players' performances are influenced, to a large extent, by the surface of the court. In other words, the surface turns out to be the pertinent latent variable for male players.  This effect is, however, less pronounced  for female players. Interestingly, we are also able to validate via our model, datasets, and algorithm that   Nadal is undoubtedly the ``King of Clay'';   Federer, a precise and accurate server, is dominant on grass (a non-clay surface other than hard court) as evidenced by his winning of Wimbledon on multiple occasions; and   Djokovic is a more ``balanced'' top player regardless of surface. Conditioned on playing on a clay court,  the probability that Nadal beats Djokovic is larger than $1/2$.  Even though the results for the women are less pronounced, our model and longitudinal dataset confirms objectively that S.~Williams, Sharapova, and Azarenka (in this order) are consistently the top three players over the past $10$ years. Such results (e.g., that Sharapova is so consistent that she is second best over  the past $10$ years) are not directly deducible from official rankings because these rankings are essentially instantaneous as they are based on a rolling $52$-week cumulative system.

\subsection{Related Work}
%To the best of our knowledge, 
Most of the works that incorporate latent factors in statistical ranking models (e.g., the BTL model) make use of mixture models. See, for example, \cite{Oh2014,NiharSimpleRobust, Suh17}. While such models are able to take into account the fact that  subpopulations within a large  population   possess different skill sets, it is difficult to make sense of what the underlying latent variable is. In contrast, by merging the BTL model with the NMF framework---the latter encouraging the  extraction of {\em parts} of complex objects---we are able to observe latent features in the learned dictionary matrix $\mathbf{W}$ (see Table~\ref{tab:temp})  and hence to extract the semantic meaning of latent variables. In our particular  application,  it is the surface type  of the court for male tennis players.   See Sec.~\ref{sec:comp} where we also show that our solution is more stable and robust (in a sense to be made precise) than that of the mixture-BTL model.  

 The paper most closely related to the present one is~\cite{ding2015} in which a topic modelling approach was used for ranking. However, unlike our work in which continuous-valued skill levels in  $\mathbf{\Lambda}$ are inferred, {\em permutations} (i.e., discrete objects) and their corresponding mixture weights were learned. We opine that our model and results provide a more {\em nuanced}  and {\em quantitative} view of the relative skill levels between players under different latent conditions. 

\subsection{Paper Outline}
In Sec.~\ref{sec:framework}, we discuss the problem setup, the  statistical model, and its associated likelihood function. In Sec.~\ref{sec:alg}, we derive   efficient  MM-based algorithms to maximize the likelihood. In Sec.~\ref{sec:expt}, we  discuss   numerical results of extensive experiments on real-world tennis datasets. We conclude our discussion in Sec.~\ref{sec:con}.
\section{Problem Setup, Statistical Model, and Likelihood }\label{sec:framework}
\subsection{Problem Definition and Model} \label{PD}
Given $N$ players and $M$ tournaments over a fixed number of years (in our case, this is  $10$), we consider a  dataset $\mathcal{D} :=  \big\{ b_{ij}^{(m)} \in\{0,1,2,\ldots\} : (i,j) \in \cP_{m} \big\}_{m=1}^M$, where $\cP_{m}$ denotes the set of games between pairs of players that  have played at least once in tournament $m$, and $b_{ij}^{(m)}$ is the number of times that player $i$ has beaten player $j$ in tournament $m$ over the fixed number of year.

To model the  skill levels of each player, we consider  a nonnegative matrix $\mathbf{\Lambda}$ of dimensions $M \times N$. The $(m,i)^{\text{th}}$ element $[\mathbf{\Lambda}]_{mi}$   represents the skill level of player $i$ in tournament $m$. Our goal is to design an algorithm to find a factorization of $\mathbf{\Lambda}$ into two  nonnegative matrices $\mathbf{W} \in\mathbb{R}_+^{M\times K}$ and $\mathbf{H}\in\mathbb{R}_+^{K\times N}$ such that the likelihood of $\mathcal{D}$ under the BTL model   in~\eqref{eqn:btl} is maximized; this is the so-called {\em maximum likelihood framework}. Here $K\le\min\{M,N\}$ is a small integer so the factorization is low-rank. In Sec.~\ref{sec:norm}, we   discuss different strategies to normalize $\mathbf{W}$ and $\mathbf{H}$ so that they are easily interpretable, e.g., as probabilities.  %In the context of  NMF, $\mathbf{W}$ and $\mathbf{H}$  are respectively known as the {\em dictionary} and {\em coefficient matrices} respectively. Their sizes are  $M \times K$ and $K \times N$ respectively where $K \le \min\{M,N\}$ is a small integer. 
 Roughly speaking, the eventual interpretation of $\mathbf{W}$ and $\mathbf{H}$ is as follows. Each column of the dictionary matrix $\mathbf{W}$ encodes the ``likelihood'' that a certain tournament $m \in \{1,\ldots, M\}$ belongs to a certain latent class (e.g., type of surface). Each row of the coefficient matrix encodes the player's skill level in a tournament of a certain latent class. For example, referring to Fig.~\ref{fig:nmf_btl}, if the latent classes indeed correspond to surface types,  the $(1,1)$ entry of $\mathbf{W}$ could represent the likelihood that Wimbledon is  a tournament that is played on clay. The $( 1,1)$ entry of $\mathbf{H}$ could represent  Nadal's skill level on clay.   % We anticipate that values of $\mathbf{W}$ and $\mathbf{H}$ should indicate some latent variables that have physical interpretation.

\subsection{Likelihood of the BTL-NMF Model}

According to the BTL model and the notations above, the probability that player $i$ beats player $j$ in tournament $m$ is 
\begin{equation*}
    \Pr(i  \;\text{beats}\; j \; \text{in tournament}\; m) =    \frac{[\mathbf{\Lambda}]_{mi}}{[\mathbf{\Lambda}]_{mi} + [\mathbf{\Lambda}]_{mj}}.
\end{equation*}
 We expect that $\mathbf{\Lambda}$ is close to a low-rank matrix as the number of latent factors governing players' skill levels is small. We would like to exploit the ``mutual information'' or ``correlation'' between tournaments of similar characteristics to 
%As the goal is to 
find a factorization of  $\mathbf{\Lambda}$. If $\mathbf{\Lambda}$ were unstructured, we could solve $M$ independent, tournament-specific  problems to learn $(\bm{\lambda}_1,\ldots,\bm{\lambda}_M)$.  We replace $\mathbf{\Lambda}$ by $\mathbf{W}\mathbf{H}$ and the   {\em likelihood}  over all games in all tournaments (i.e., of the dataset $\mathcal{D}$), assuming conditional independence across tournaments  and games, is
\begin{equation*}
    p(\mathcal{D}|\mathbf{W},\mathbf{H})=\prod\limits_{m=1}^{M}\prod\limits_{(i,j) \in \cP_{m}} \bigg( \frac{[\mathbf{W}\mathbf{H}]_{mi}}{[\mathbf{W}\mathbf{H}]_{mi} + [\mathbf{W}\mathbf{H}]_{mj}}\bigg)^{b_{ij}^{(m)}}.
\end{equation*}
It is often more tractable to minimize the {\em    negative log-likelihood}. In the sequel, we regard this as our objective function which can be expressed as
\begin{align}
&    f( \mathbf{W},\mathbf{H}) := -\log p(\mathcal{D}|\mathbf{W},\mathbf{H})  \nonumber\\
    &\;\;= \sum\limits_{m=1}^{M}\sum\limits_{(i,j) \in \cP_{m}} b_{ij}^{(m)}\Big[ -\log\big([\mathbf{W}\mathbf{H}]_{mi}\big) +\log\big([\mathbf{W}\mathbf{H}]_{mi} + [\mathbf{W}\mathbf{H}]_{mj}\big) \Big]. \label{eqn:neg_ll}
\end{align}

\section{Algorithms and Theoretical Guarantees} \label{sec:alg}
In this section, we describe the algorithm to optimize \eqref{eqn:neg_ll}, together with accompanying theoretical guarantees. We also discuss how we ameliorate numerical problems while maintaining the desirable guarantees of the algorithm.
\subsection{Majorization-Minimization (MM) Algorithm} \label{sec:MM}
We now describe how we use an MM algorithm~\cite{hunterLange2004} to optimize~\eqref{eqn:neg_ll}. The MM framework iteratively solves the problem of minimizing a certain function $f(x)$, but its utility is most evident when  the direct of optimization of $f(x)$ is difficult. One proposes an {\em auxiliary function} or {\em  majorizer} $u(x,x')$ that satisfies the following two properties: (i) $f(x) = u(x,x) ,\forall\, x$ and (ii) $f(x)\le u(x,x'),\forall\, x,x'$ (majorization). In addition for a fixed value of $x'$, the minimization of $u(\cdot, x')$ is assumed to be tractable (e.g., there exists a closed-form solution for $x^*=\argmin_x u(x,x')$). Then, one adopts an iterative approach to find a sequence $\{ x^{(l)} \}_{l =1}^\infty$. One observes that if $x^{ (l+1) } = \argmin_{x} u(x,x^{(l )})$ is a minimizer at iteration $l+1$, then 
\begin{equation}
f(x^{(l+1)}) \stackrel{ \text{(ii)} }{\le} u(x^{(l+1)} , x^{(l)})  \le u(x^{(l )} , x^{(l)})  \stackrel{\text{(i)}}{=} f(x^{(l)}).\label{eqn:mm}
\end{equation}
Hence, if such an auxiliary function $u(x,x')$ can be found, it is guaranteed that the sequence of iterates results in a sequence of non-increasing objective values. 

Applying MM to our model is slightly more involved as we are trying to find {\em two} nonnegative matrices $\mathbf{W} $ and $\mathbf{H}$. Borrowing ideas from using MM in NMFs problems (see for example the works~\cite{fevotte2011algorithms,tan2013automatic}), the procedure first  updates $\mathbf{W}$ by keeping $\mathbf{H}$  fixed,   then updates $\mathbf{H}$ by keeping $\mathbf{W}$ fixed to its previously updated value. We will describe, in the following, how to  optimize the original objective in \eqref{eqn:neg_ll} with respect to $\mathbf{W}$ with fixed $\mathbf{H}$ as the other optimization proceeds in an almost\footnote{The updates for $\mathbf{W}$ and $\mathbf{H}$ are not completely symmetric because the data is in the form of a $3$-way tensor  $\{b_{ij}^{(m)}\}$; this is also apparent in the objective in~\eqref{eqn:neg_ll} and the updates in~\eqref{eqn:ori_update}.}   symmetric fashion since $\mathbf{\Lambda}^T=\mathbf{H}^T\mathbf{W}^T$.  As mentioned above, the MM algorithm requires us to construct  an auxiliary function $u_{1}(\mathbf{W},\tilde{\mathbf{W}}|\mathbf{H})$ that majorizes $-\log p(\mathcal{D}|\mathbf{W}, \mathbf{H})$.

The difficulty in optimizing the original objective function in \eqref{eqn:neg_ll} is twofold. The first concerns  the coupling of the two terms $[\mathbf{W}\mathbf{H}]_{mi}$ and $[\mathbf{W}\mathbf{H}]_{mj}$ inside the logarithm function. We resolve this using a technique introduced by Hunter in~\cite{hunter2004mm}. It is known that for any concave function $f$, its first-order Taylor approximation overestimates it, i.e.,  $f(y) \leq f(x) + \nabla f(x)^{T}(y-x)$. Since the  logarithm function is concave, we have the inequality $\log y \leq \log x + \frac{1}{x}(y-x)$ which is an equality when $x=y$. These two properties mean that  the following  is a majorizer of the term $\log ([\mathbf{W}\mathbf{H}]_{mi} + [\mathbf{W}\mathbf{H}]_{mj} )$ in~\eqref{eqn:neg_ll}:
\begin{equation*}
     \log \big([\mathbf{W}^{(l)}\mathbf{H}]_{mi} +  [\mathbf{W}^{(l)}\mathbf{H}]_{mj}\big) + \frac{[\mathbf{W}\mathbf{H}]_{mi} + [\mathbf{W}\mathbf{H}]_{mj}}{[\mathbf{W}^{(l)}\mathbf{H}]_{mi} +  [\mathbf{W}^{(l)}\mathbf{H}]_{mj}} -1.
\end{equation*}
The second difficulty in optimizing~\eqref{eqn:neg_ll} concerns  the term $\log ([\mathbf{W}\mathbf{H}]_{mi})=\log ( \sum_{k}w_{mk}h_{ki})$. By introducing the terms $\gamma_{mki}^{(l)} :=w_{mk}^{(l)}h_{ki}/[\mathbf{W}^{(l)}\mathbf{H}]_{mi}$ for $ k \in\{1,\ldots, K\}$ (which have the property that $\sum_k\gamma_{mki}^{(l)}=1$) to the sum in $\log ( \sum_{k}w_{mk}h_{ki})$ as was done by F\'evotte and Idier in~\cite[Theorem~1]{fevotte2011algorithms}, and using the convexity of $-\log x$   and Jensen's inequality, we obtain the following majorizer of the term $-\log ([\mathbf{W}\mathbf{H}]_{mi})$  in~\eqref{eqn:neg_ll}:
\begin{equation*}
    -\sum\limits_{k}\frac{w_{mk}^{(l)}h_{ki}}{[\mathbf{W}^{(l)}\mathbf{H}]_{mi}}\log \bigg(\frac{w_{mk}}{w_{mk}^{(l)}}[\mathbf{W}^{(l)}\mathbf{H}]_{mi}\bigg).
\end{equation*}
The same procedure can be applied to find an auxiliary function  $u_{2}(\mathbf{H},\tilde{\mathbf{H}}|\mathbf{W})$ for the optimization for $\mathbf{H}$. Minimization of the two auxiliary functions with respect to $\mathbf{W}$ and $\mathbf{H}$ leads to the following MM updates: 
{
\small
\begin{subequations} \label{eqn:ori_update}
\begin{align}
    \tilde{w}_{mk}^{(l+1)} &\leftarrow \ddfrac{\sum\limits_{(i,j) \in \cP_{m}} b_{ij}^{(m)}\frac{w_{mk}^{(l)}h_{ki}^{(l)}}{[\mathbf{W}^{(l)}\mathbf{H}^{(l)}]_{mi}}}{\sum\limits_{(i,j) \in \cP_{m}} b_{ij}^{(m)}\frac{h_{ki}^{(l)}+h_{kj}^{(l)}}{[\mathbf{W}^{(l)}\mathbf{H}^{(l)}]_{mi}+[\mathbf{W}^{(l)}\mathbf{H}^{(l)}]_{mj}}} \label{eqn:updateW},\\
    \tilde{h}_{ki}^{(l+1)} &\leftarrow \ddfrac{\sum\limits_{m} \sum\limits_{j \neq i:(i,j) \in \cP_{m}} b_{ij}^{(m)}\frac{w_{mk}^{(l+1)}h_{ki}^{(l)}}{[\mathbf{W}^{(l+1)}\mathbf{H}^{(l)}]_{mi}}}{\sum\limits_{m}\sum\limits_{j \neq i: (i,j) \in \cP_{m}} \big( b_{ij}^{(m)} + b_{ji}^{(m)} \big)\frac{w_{mk}^{(l+1)}}{[\mathbf{W}^{(l+1)}\mathbf{H}^{(l)}]_{mi} + [\mathbf{W}^{(l+1)}\mathbf{H}^{(l)}]_{mj}}}\label{eqn:updateH} .
\end{align}
\end{subequations}
}%
Note that since we first update $\mathbf{W}$, $\mathbf{H}$ is given and fixed which means that it is indexed by the previous iteration $l$; as for the update of $\mathbf{H}$, the newly calculated $\mathbf{W}$ at iteration $l+1$ will be used. 
\subsection{Resolution of Numerical Problems}\label{sec:num}
While the above updates   guarantee that the objective function does not decrease, numerical problems may arise in the  implementation of~\eqref{eqn:ori_update}. Indeed, it is possible that $[\mathbf{W}\mathbf{H}]_{mi}$ becomes extremely close to zero for some $(m,i)$. 
%During the running of the algorithm on some dataset which will be elaborated further in the experiment part, we are faced with some numerical problem: division by zero. By looking into the location of the problem in the matrix, we notice it is possible that $[\mathbf{W}\mathbf{H}]_{mi}$ becomes extremely close to zero for some $(m,i)$. For example, in one of the initialization with $K=2$, $\underline{w}_{m}^{(l)} = (1.56e-292, 1.86e-0.1)$, and $\mathbf{h}_{i}^{(l)} = (3.13e-02, 0.00e+00)^{T}$, the inner product leads to $4.88e-294$, and this extremely small number might raise error as it appears in the denominator. 
To prevent such numerical problems from arising, our strategy is to add a small number $\epsilon>0$ to every element of $\mathbf{H}$ in~\eqref{eqn:neg_ll}. The intuitive explanation that justifies this is that we believe that each player has some default skill level in every type of tournament. By modifying $\mathbf{H}$ to $\mathbf{H}+\epsilon\mathds{1}$, where $\mathds{1}$ is the $K \times N$ all-ones matrix, we have the following new objective function:
\begin{align}
  f_\epsilon(\mathbf{W},\mathbf{H}) &:=
    \sum\limits_{m=1}^{M}\sum\limits_{(i,j) \in \cP_{m}} b_{ij}^{(m)} \Big[ - \log\big([\mathbf{W}(\mathbf{H}+\epsilon\mathds{1})]_{mi}\big)\nonumber\\
&\qquad\qquad     + \log\big([\mathbf{W}(\mathbf{H}+\epsilon\mathds{1})]_{mi} + [\mathbf{W}(\mathbf{H}+\epsilon\mathds{1})]_{mj}\big) \Big]\label{eqn:new_ll_func} .
\end{align}
Note that $f_0(\mathbf{W},\mathbf{H})=f(\mathbf{W},\mathbf{H})$, defined in \eqref{eqn:neg_ll}.   Using the same ideas involving MM to optimize $f(\mathbf{W},\mathbf{H})$ as in Sec.~\ref{sec:MM}, we can find new auxiliary functions,  denoted similarly as $u_{1}(\mathbf{W},\tilde{\mathbf{W}}|\mathbf{H})$  and $u_{2}( \mathbf{H} , \tilde{\mathbf{H}}|\mathbf{W})$,  leading to following updates
{
\small
\begin{subequations} \label{eqn:new_update}
\begin{align}
    \tilde{w}_{mk}^{(l+1)} &\leftarrow \ddfrac{\sum\limits_{(i,j) \in\cP_m} b_{ij}^{(m)} \frac{w_{mk}^{(l)}(h_{ki}^{(l)}+\epsilon)}{[\mathbf{W}^{(l)}(\mathbf{H}^{(l)}+\epsilon\mathds{1})]_{mi}}}{\sum\limits_{(i,j)\in\cP_m} b_{ij}^{(m)} \frac{h_{ki}^{(l)} + h_{kj}^{(l)} + 2\epsilon}{[\mathbf{W}^{(l)}(\mathbf{H}^{(l)}+\epsilon\mathds{1})]_{mi} + [\mathbf{W}^{(l)}(\mathbf{H}^{(l)}+\epsilon\mathds{1})]_{mj}}}\label{eqn:updateW_new},\\
    \tilde{h}_{ki}^{(l+1)} &\leftarrow\ddfrac{\sum\limits_{m}\sum\limits_{j \neq i: (i,j) \in \cP_{m}} b_{ij}^{(m)} \frac{w_{mk}^{(l+1)}(h_{ki}^{(l)}+\epsilon)}{[\mathbf{W}^{(l+1)}(\mathbf{H}^{(l)}+\epsilon\mathds{1})]_{mi}}}{\sum\limits_{m}\sum\limits_{j \neq i: (i,j) \in \cP_{m}}  \frac{(b_{ij}^{(m)} + b_{ji}^{(m)})w_{mk}^{(l+1)}}{[\mathbf{W}^{(l+1)}(\mathbf{H}^{(l)}+\epsilon\mathds{1})]_{mi} + [\mathbf{W}^{(l+1)}(\mathbf{H}^{(l)}+\epsilon\mathds{1})]_{mj}}} - \epsilon.\label{eqn:updateH_new}
\end{align} \end{subequations}
}%
 Notice that although this solution successfully prevents division by zero (or small numbers) during the iterative process, for the new update of $\mathbf{H}$, it is possible $h_{ki}^{(l+1)}$ becomes negative because of the subtraction by $\epsilon$ in \eqref{eqn:updateH_new}. To ensure  $h_{ki}$ is nonnegative as required by the nonnegativity of NMF, we set
\begin{equation}
    \tilde{h}_{ki}^{(l+1)} \leftarrow \max \big\{\tilde{h}_{ki}^{(l+1)}, 0\big\}. \label{eqn:trunc}
\end{equation}
%This means that if $\tilde{h}_{ki}^{(l+1)}<0$, we set it to be $0$. 
 After this truncation operation, it is, however, unclear whether the likelihood function is non-decreasing, as we have altered the vanilla MM procedure. 
 
 We now prove that $f_\epsilon$ in  \eqref{eqn:new_ll_func} is non-increasing as the iteration count increases.  Suppose for the $(l+1)^{\text{st}}$ iteration for  $\tilde{\mathbf{H}}^{(l+1)}$, truncation to zero only occurs for the  $(k,i)^{\text{th}}$ element and  and all other elements stay unchanged, meaning $\tilde{h}_{ki}^{(l+1)} = 0$ and $\tilde{h}_{k', i'}^{(l+1)} = \tilde{h}_{k' ,i ' }^{(l)}$ for all $(k',i') \neq (k, i)$. We would like to show that $f_\epsilon(\mathbf{W},\tilde{\mathbf{H}}^{(l+1)}) \leq f_\epsilon( \mathbf{W},\tilde{\mathbf{H}}^{(l)})$. It suffices to show $u_{2}(\tilde{\mathbf{H}}^{(l+1)}, \tilde{\mathbf{H}}^{(l)}|\mathbf{W}) \leq f_\epsilon(\mathbf{W},\tilde{\mathbf{H}}^{(l)})$, because if this is true, we have the following inequality
\begin{equation}
    f_\epsilon(\mathbf{W},\tilde{\mathbf{H}}^{(l+1)}) 
    \leq u_{2}(\tilde{\mathbf{H}}^{(l+1)}, \tilde{\mathbf{H}}^{(l)}|\mathbf{W}) \leq f_\epsilon(\mathbf{W},\tilde{\mathbf{H}}^{(l)}), \label{eqn:follow_ineq}
\end{equation}
 where the first inequality holds as $u_2$ is an auxiliary function for  $\mathbf{H}$. The truncation is invoked only when the update in~\eqref{eqn:updateH_new} becomes negative, i.e., when
\begin{equation*}
    \frac{\sum\limits_{m}\sum\limits_{j \neq i:(i,j) \in \cP_{m}} b_{ij}^{(m)} \frac{w_{mk}^{(l+1)}(h_{ki}^{(l)}+\epsilon)}{[\mathbf{W}^{(l+1)}(\mathbf{H}^{(l)}+\epsilon\mathds{1})]_{mi}}}{\sum\limits_{m}\sum\limits_{j \neq i:(i,j) \in \cP_{m}}\frac{ (b_{ij}^{(m)} + b_{ji}^{(m)}) w_{mk}^{(l+1)}}{[\mathbf{W}^{(l+1)}(\mathbf{H}^{(l)}+\epsilon\mathds{1})]_{mi} + [\mathbf{W}^{(l+1)}(\mathbf{H}^{(l)}+\epsilon\mathds{1})]_{mj}}} \leq \epsilon.
\end{equation*}
Using this inequality and performing some straightforward but tedious algebra as shown in Sec. S-1 in the supplementary material~\cite{xia2019}, we can justify the second inequality in~\eqref{eqn:follow_ineq} as follows
{\small\begin{align*} 
&f_\epsilon(\mathbf{W},\tilde{\mathbf{H}}^{(l)}) - u_{2}(\tilde{\mathbf{H}}^{(l+1)}, \tilde{\mathbf{H}}^{(l)}|\mathbf{W}) \nonumber\\
&\geq    \sum\limits_{m}\sum\limits_{j \neq i :(i,j)  \in \cP_{m}} \frac{(b_{ij}^{(m)}+b_{ji}^{(m)})w_{mk}}{[\mathbf{W}(\mathbf{H}^{(l)}\!+\!\epsilon\mathds{1})]_{mi} \!+\! [\mathbf{W}(\mathbf{H}^{(l)}\!+\!\epsilon\mathds{1})]_{mj}}  \bigg[h_{ki}^{(l)} \!-\!\epsilon \log\Big(\frac{h_{ki}^{(l)}\!+\!\epsilon}{\epsilon}\Big)  \bigg] \!\geq\! 0.
\end{align*}}%
The last inequality follows because  $b_{ij}^{(m)}$,  $\mathbf{W}$ and $\mathbf{H}^{(l)}$ are nonnegative, and $h_{ki}^{(l)}-\epsilon \log(\frac{h_{ki}^{(l)}+\epsilon}{\epsilon}) \geq 0$  since $x \geq \log(x+1)$ for all $x \geq 0$ with equality at $x=0$. Hence, the likelihood is non-decreasing during the MM update even though we included an additional operation that truncates $\tilde{h}_{ki}^{(l+1)}<0$ to zero in \eqref{eqn:trunc}.

\subsection{Normalization} \label{sec:norm}
%Since NMF helps to estimate the hidden components with specific physical interpretation, we wish to put some constraints on the value of the element of $\mathbf{W}$ and $\mathbf{H}$, so that the values can be interpreted as probabilities over some latent variables. It is crucial that we keep the likelihood unchanged after normalization, which is keeping $\frac{[\mathbf{W}\mathbf{H}]_{mi}}{[\mathbf{W}\mathbf{H}]_{mi}+[\mathbf{W}\mathbf{H}]_{mj}}$ unchanged for $\forall (m,i,j)$. As for $\mathbf{H}$, since there are $mi$ and $mj$ terms appearing in the denominator but only $mi$ in the numerator, the idea is to normalize over all elements of $\mathbf{H}$ to keep the fraction same. As for $\mathbf{W}$, since only $m$ term appear both in numerator and denominator, we can impose either row or column normalization.

It is well-known that NMF is not unique in the general case, and it is characterized by a scale and permutation indeterminacies~\cite{CichockiBook}. For the problem at hand, for the learned $\mathbf{W}$ and $\mathbf{H}$ matrices to be interpretable as ``skill levels'' with respect to different latent variables, it is imperative we consider {\em normalizing} them appropriately after every MM iteration in \eqref{eqn:new_update}. However, there are different ways to normalize the entries in the matrices and one has to ensure that after normalization, the likelihood of the model stays unchanged. This is tantamount to keeping the ratio $\frac{[\mathbf{W}(\mathbf{H}+\epsilon\mathds{1}) ]_{mi}}{[\mathbf{W}(\mathbf{H}+\epsilon\mathds{1})]_{mi}+[\mathbf{W}(\mathbf{H}+\epsilon\mathds{1})]_{mj}}$  unchanged   for all  $(m,i,j)$. The key observations here are twofold: First, concerning $\mathbf{H}$, since terms indexed by $(m,i)$ and $(m,j)$ appear in the  denominator but only $(m,i)$ appears  in the numerator, we can normalize over all elements of $\mathbf{H}$ to keep this  fraction unchanged. Second, concerning $\mathbf{W}$, since only terms indexed by $m$ term appear both in numerator and denominator, we can normalize either rows or columns as we will show in the following. 

\subsubsection{Row Normalization of $\mathbf{W}$ and Global Normalization of $\mathbf{H}$}\label{sec:row_norm}
Define the row sums of $\mathbf{W}$ as $r_{m} := \sum_{k}\tilde{w}_{mk}$ and let $\alpha := \frac{\sum_{k,i}\tilde{h}_{ki}+KN\epsilon}{1+KN\epsilon}$. Now consider the following operations:
\begin{equation*}
    w_{mk} \leftarrow \frac{\tilde{w}_{mk}}{r_{m}},\quad\mbox{and}\quad  h_{ki}\leftarrow \frac{\tilde{h}_{ki}+(1-\alpha)\epsilon}{\alpha}.
\end{equation*}
The above update to obtain $h_{ki}$ may result in it being negative; however, the truncation operation in~\eqref{eqn:trunc} ensures that  $h_{ki}$  is eventually nonnegative.\footnote{One might be tempted to normalize $\mathbf{H}+\epsilon\mathds{1}\in\mathbb{R}_{+}^{K\times N}$. This, however, does not resolve   numerical issues (analogous to division by zero in~\eqref{eqn:ori_update}) as some entries of $\mathbf{H}+\epsilon\mathds{1}$ may be zero. }   See also the update to obtain $\tilde{h}_{ki}^{(l+1)}$ in Algorithm~\ref{alg:btl_nmf}.
The operations above  keep the likelihood unchanged and achieve the desired row normalization of $\mathbf{W}$  since
{
\small
\begin{align*}
    & \frac{\sum_{k} \tilde{w}_{mk} (\tilde{h}_{ki} + \epsilon)}{\sum_{k} \tilde{w}_{mk} (\tilde{h}_{ki} + \epsilon) + \sum_{k} \tilde{w}_{mk} (\tilde{h}_{kj} + \epsilon)}
    = \frac{\sum_{k} \frac{\tilde{w}_{mk}}{r_{m}} (\tilde{h}_{ki} + \epsilon)}{\sum_{k} \frac{\tilde{w}_{mk}}{r_{m}} (\tilde{h}_{ki} + \epsilon) + \sum_{k} \frac{\tilde{w}_{mk}}{r_{m}} (\tilde{h}_{kj} + \epsilon)}\\
    &=  \frac{\sum_{k} w_{mk} \frac{(\tilde{h}_{ki} + \epsilon)}{\alpha}}{\sum_{k} w_{mk} \frac{(\tilde{h}_{ki} + \epsilon)}{\alpha} + \sum_{k} w_{mk} \frac{(\tilde{h}_{ki} + \epsilon)}{\alpha}} = \frac{\sum_{k} w_{mk} (h_{ki} + \epsilon)}{\sum_{k} w_{mk} (h_{ki} + \epsilon) + \sum_{k} w_{mk} (h_{kj} + \epsilon)}.
\end{align*} 
}%

\begin{algorithm}[t]
    \caption{MM Alg.\ for  BTL-NMF model with column normalization of~$\mathbf{W}$}
    \begin{algorithmic}
    \STATE \textbf{Input:} $M$ tournaments; $N$ players; number of times player $i$ beats player $j$ in tournament $m$ in dataset  
    $\mathcal{D} = \big\{ b_{ij}^{(m)}: i,j \in \{1,...,N\}, m \in \{1,...,M\}\big\}$
    \STATE \textbf{Init:} Fix $K \in\mathbb{N}$, $\epsilon>0$, $\tau>0$ and initialize $\mathbf{W}^{(0)} \in \mathbb{R}_{++}^{M \times K}, \mathbf{H}^{(0)} \in \mathbb{R}_{++}^{K \times N}$.
    \WHILE {diff $\geq \tau>0$}
    
    \STATE 
    \begin{enumerate}[(1)]
        \item \textbf{Update} $\forall m \in \{1,...,M\}, \forall k \in \{1,...,K\}, \forall i \in \{1,...,N\}$\\
        $\tilde{w}_{mk}^{(l+1)} = \frac{\sum\limits_{ i,j } b_{ij}^{(m)} \frac{w_{mk}^{(l)}(h_{ki}^{(l)}+\epsilon)}{[\mathbf{W}^{(l)}(\mathbf{H}^{(l)}+\epsilon\mathds{1})]_{mi}}}{\sum\limits_{ i,j } b_{ij}^{(m)} \frac{h_{ki}^{(l)} + h_{kj}^{(l)} + 2\epsilon}{[\mathbf{W}^{(l)}(\mathbf{H}^{(l)}+\epsilon\mathds{1})]_{mi} + [\mathbf{W}^{(l)}(\mathbf{H}^{(l)}+\epsilon\mathds{1})]_{mj}}}$\\
        $\tilde{h}_{ki}^{(l+1)} = \max \Bigg\{ \frac{\sum\limits_{m}\sum\limits_{j \neq i} b_{ij}^{(m)} \frac{w_{mk}^{(l+1)}(h_{ki}^{(l)}+\epsilon)}{[\mathbf{W}^{(l+1)}(\mathbf{H}^{(l)}+\epsilon\mathds{1})]_{mi}}}{\sum\limits_{m}\sum\limits_{j \neq i} \frac{(b_{ij}^{(m)} + b_{ji}^{(m)}) w_{mk}^{(l+1)}}{[\mathbf{W}^{(l+1)}(\mathbf{H}^{(l)}+\epsilon\mathds{1})]_{mi} + [\mathbf{W}^{(l+1)}(\mathbf{H}^{(l)}+\epsilon\mathds{1})]_{mj}}} - \epsilon,0\Bigg\}$
        \item \textbf{Normalize} $\forall\, m \in \{1,...,M\}, \forall\, k \in \{1,...,K\}, \forall\, i \in \{1,...,N\}$\\
        $w_{mk}^{(l+1)} \leftarrow \frac{\tilde{w}_{mk}^{(l+1)}}{\sum\limits_{m}\tilde{w}_{mk}^{(l+1)}}$; \ 
        $\hat{h}_{ki}^{(l+1)} \leftarrow \tilde{h}_{ki}^{(l+1)}\sum\limits_{m}\tilde{w}_{mk}^{(l+1)} + \epsilon\Big(\sum\limits_{m}\tilde{w}_{mk}^{(l+1)}-1\Big)$\\
        Calculate $\beta = \frac{\sum_{k,i}\hat{h}_{ki}^{(l+1)}+KN\epsilon}{1+KN\epsilon}$, $h_{ki}^{(l+1)} \leftarrow \frac{\hat{h}_{ki}^{(l+1)}+(1-\beta)\epsilon}{\beta}$
        \item diff $\leftarrow \max\Big\{\max\limits_{ m,k }\big|w_{mk}^{(l+1)}-w_{mk}^{(l)}\big|,\max\limits_{ k,i }\big|h_{ki}^{(l+1)}-h_{ki}^{(l)}\big|\Big\}$
    \end{enumerate}
    \ENDWHILE
    \RETURN $(\mathbf{W}, \mathbf{H})$ that forms a local maximizer of the likelihood $ p(\mathcal{D}|\mathbf{W},\mathbf{H})$
\end{algorithmic}
\label{alg:btl_nmf}
\end{algorithm} 

\subsubsection{Column Normalization of $\mathbf{W}$ and Global Normalization of $\mathbf{H}$}\label{sec:col_norm}
Define the column sums of $\mathbf{W}$ as $c_{k} := \sum_{m}\tilde{w}_{mk}$ and let $\beta := \frac{\sum_{k,i}\hat{h}_{ki}+KN\epsilon}{1+KN\epsilon}$. Now consider the following operations:
\begin{equation*} 
    w_{mk} \leftarrow \frac{\tilde{w}_{mk}}{c_{k}},\quad \hat{h}_{ki}\leftarrow \tilde{h}_{ki}c_{k} + \epsilon(c_{k}-1),\quad\mbox{and}\quad    h_{ki}\leftarrow \frac{\hat{h}_{ki}+(1-\beta)\epsilon}{\beta}. 
\end{equation*}
This would keep the likelihood unchanged and achieve the desired column normalization of $\mathbf{W}$ since 
{\small \begin{align*}
   & \frac{\sum_{k} \tilde{w}_{mk} (\tilde{h}_{ki} + \epsilon)}{\sum_{k} \tilde{w}_{mk} (\tilde{h}_{ki} + \epsilon) + \sum_{k} \tilde{w}_{mk} (\tilde{h}_{kj} + \epsilon)}
    = \frac{\sum_{k} \frac{\tilde{w}_{mk}}{c_{k}} (\tilde{h}_{ki} + \epsilon)c_{k}}{\sum_{k} \frac{\tilde{w}_{mk}}{c_{k}} (\tilde{h}_{ki} + \epsilon)c_{k} + \sum_{k} \frac{\tilde{w}_{mk}}{c_k} (\tilde{h}_{kj} + \epsilon)c_{k}}\\
    &= \frac{\sum_{k} w_{mk} \frac{(\hat{h}_{ki} + \epsilon)}{\beta}}{\sum_{k} w_{mk} \frac{(\hat{h}_{ki} + \epsilon)}{\beta} + \sum_{k} w_{mk} \frac{(\hat{h}_{ki} + \epsilon)}{\beta}} = \frac{\sum_{k} w_{mk} (h_{ki} + \epsilon)}{\sum_{k} w_{mk} (h_{ki} + \epsilon) + \sum_{k} w_{mk} (h_{kj} + \epsilon)}.\vspace{-.1in}
\end{align*}}
\noindent Using this normalization strategy, it is easy to verify that all the entries of $\mathbf{\Lambda} = \mathbf{W}\mathbf{H}$ sum to one.\footnote{We have $\sum_{m}\sum_{i}[\mathbf{\Lambda}]_{mi} = \sum_{m}\sum_{i}\sum_{k}w_{mk}h_{ki} = \sum_{i}\sum_{k}h_{ki}\!\!\sum_{m}\!\!w_{mk} = \sum_{k,i}\!\!h_{ki}= 1$.} This allows us to interpret the entries of $\mathbf{\Lambda}$ as ``conditional probabilities''.
%\begin{equation*}
%	\sum_{m}\sum_{i}[\Lambda]_{mi} = \sum_{m}\sum_{i}\sum_{k}w_{mk}h_{ki} = \sum_{i}\sum_{k}h_{ki}\sum_{m}w_{mk} = \sum_{k,i}h_{ki}= 1
%\end{equation*}

\subsection{Summary of Algorithm}
Algorithm \ref{alg:btl_nmf} presents pseudo-code for  optimizing \eqref{eqn:new_ll_func} with columns of $\mathbf{W}$ normalized. The algorithm when the rows of $\mathbf{W}$ are normalized is similar; we replace the normalization step with the procedure outlined in Sec.~\ref{sec:row_norm}.  In summary, we have proved that the sequence of iterates $\{ (\mathbf{W}^{(l )} , \mathbf{H}^{(l )} ) \}_{l =1}^\infty$ results in the sequence of objective functions $\{f_\epsilon(\mathbf{W}^{(l )} , \mathbf{H}^{(l )} ) \}_{l =1}^\infty$ being non-increasing. Furthermore, if $\epsilon>0$,   numerical problems do not arise and  with the normalization as described in Sec.~\ref{sec:col_norm},   %the columns of $\mathbf{W}$ and the entire matrix $\mathbf{H}$ are normalized to $1$ so 
the entries  in $\mathbf{\Lambda}$ can be interpreted as ``conditional probabilities'' as we will further illustrate in Sec.~\ref{sec:men}.

\subsection{Convergence of Matrices $\{(\mathbf{W}^{(l )}, \mathbf{H}^{(l )})\}_{l=1}^\infty$ to Stationary Points}
While we have proved that the sequence of objectives $\{f_\epsilon(\mathbf{W}^{(l )} , \mathbf{H}^{(l )} ) \}_{l =1}^\infty$ is non-increasing (and hence it converges because  it is bounded), it is not clear as to whether the sequence of {\em iterates} generated by the algorithm $\{ (\mathbf{W}^{(l  )} , \mathbf{H}^{(l )} ) \}_{l =1}^\infty$  converges and if so to what. We define the {\em marginal functions} $f_{1,\epsilon}(\mathbf{W} | \overline{\mathbf{H}}) := f_\epsilon(\mathbf{W}, \overline{\mathbf{H}})$ and $f_{2,\epsilon} (\mathbf{H}| \overline{\mathbf{W}}) := f_\epsilon(\overline{\mathbf{W}}, \mathbf{H})$. For any function $g:\mathcal{D}\to\mathbb{R}$, we let $g'( x; d) := \liminf_{\lambda\downarrow 0} (g( x+\lambda d)-g(x))/\lambda$ be the {\em directional derivative} of $g$ at point $x$ in direction $d$.  We say that $(\overline{\mathbf{W}}, \overline{\mathbf{H}})$ is a {\em stationary point} of the minimization problem 
\begin{equation}
\min_{ \mathbf{W}  \in\mathbb{R}_{+}^{M\times K},  \mathbf{H}  \in\mathbb{R}_{+}^{K\times N}} f_\epsilon (\mathbf{W},\mathbf{H}) \label{eqn:min}
\end{equation}
 if  the following two conditions hold:
 \begin{alignat*}{2}
  f_{1,\epsilon}'(\overline{\mathbf{W}};\mathbf{W} - \overline{\mathbf{W}}  | \overline{\mathbf{H}}) &\ge 0, \qquad\forall\,& \mathbf{W}&\,\in\mathbb{R}_+^{M\times K},\\
  f_{2,\epsilon}'(\overline{\mathbf{H}};\mathbf{H} - \overline{\mathbf{H}}  | \overline{\mathbf{W}}) &\ge 0, \qquad\forall\,& \mathbf{H}&\in\mathbb{R}_+^{K\times N}.
\end{alignat*}
%\begin{align}
%  f_{1,\epsilon}'(\overline{\mathbf{W}};\mathbf{W} - \overline{\mathbf{W}}  | \overline{\mathbf{H}}) \ge 0,\quad   f_{2,\epsilon}'(\overline{\mathbf{H}};\mathbf{H} - \overline{\mathbf{H}}  | \overline{\mathbf{W}}) &\ge 0 \quad \forall (\mathbf{W},\mathbf{H}) \in \mathbb{R}_+^{M\times K}\times \mathbb{R}_+^{K\times N}
%\end{align}
This definition generalizes the usual notion of a stationary point when the function  is differentiable and the domain is unconstrained (i.e., $\overline{x}$ is a stationary point if $\nabla f (\overline{x})=0$). However, in our NMF setting, the matrices are constrained to be nonnegative, hence the need for this generalized definition.

If the matrices are initialized to some $\mathbf{W}^{(0)}$ and $\mathbf{H}^{(0)}$  that are (strictly) positive and $\epsilon>0$, then we have the following desirable property. 

%If we want to further draw a convergence conclusion of the update, we need more constraints on the initialization. Instead of initializing $\mathbf{W}^{(0)}$ and $\mathbf{H}^{(0)}$ to nonnegative values, we need both matrices to have positive entries.
\begin{theorem} \label{thm:conv}
	If $\mathbf{W}$ and $ \mathbf{H}$ are initialized to have positive entries (i.e., $\mathbf{W}^{(0)} \in \mathbb{R}^{M \times K}_{++} = (0,\infty)^{M \times K}$ and $\mathbf{H}^{(0)} \in \mathbb{R}^{K \times N}_{++}$) and $\epsilon>0$, then every limit point of $\{(\mathbf{W}^{(l )}, \mathbf{H}^{(l )})\}_{l=1}^{\infty}$ generated by Algorithm \ref{alg:btl_nmf} is a  stationary point  of~\eqref{eqn:min}.
\end{theorem}

Thus, apart from ensuring that there are no numerical errors, another reason why we incorporate $\epsilon>0$ in the modified objective function in~\eqref{eqn:new_ll_func} is because a stronger convergence guarantee can be ensured.  The proof of this theorem, provided in Sec. S-2 of~\cite{xia2019}, follows along the lines of the main result in Zhao and Tan~\cite{zhao2017unified}, which itself hinges on the convergence analysis of block successive minimization methods provided by Razaviyayn, Hong, and Luo~\cite{Raz}. In essence, we need to verify that $f_{1,\epsilon}$ and $f_{2,\epsilon}$ together with their auxiliary functions $u_1$ and $u_2$ satisfy the five regularity conditions in Definition 3 of~\cite{zhao2017unified}. However, there are some important differences vis-\`a-vis~\cite{zhao2017unified} (e.g., analysis of the normalization step in Algorithm~\ref{alg:btl_nmf}) which   we describe in detail in Remark~1 of~\cite{xia2019}.  %The  proof of Theorem~\ref{thm:conv} is provided in Sec.~\ref{doc-sec:conv} of the  supplementary material~\cite{xia2019}.
\section{Numerical Experiments and Discussion}\label{sec:expt}
In this section, we describe how the datasets are collected and provide interesting and insightful interpretations of the numerical  results. All datasets and code can be found at the following GitHub repository~\cite{xia2019}.

%\begin{changemargin}{-10mm}{-10mm}
\begin{table}[t]
\centering
\caption{Partial men's dataset for the French Open}\label{tab:partial}
\small
 \begin{tabular}{|c||c|c|c|c|} 
 \hline
 Against & R. Nadal & N. Djokovic & R. Federer & A. Murray \\ 
 \hline
 R. Nadal &0&5&3&2\\
 \hline
 N. Djokovic & 1&0&1&2\\
 \hline
 R. Federer & 0&1&0&0\\
 \hline
 A. Murray & 0&0&0&0\\
 \hline
\end{tabular}
\end{table}
%\end{changemargin}

\begin{table}[t]
\centering
\caption{Sparsity of datasets $\{ b_{ij}^{(m)} \}$}\label{tab:sparsity}
	\begin{tabular}{|c||c|c|c|c|}
		\hline
		 & \multicolumn{2}{c|}{ \textbf{Male}} &\multicolumn{2}{c|}{ \textbf{Female}}\\
		 \hline
		 \textbf{Total Entries} & \multicolumn{2}{c|}{$14\times 20\times 20=5600$} & \multicolumn{2}{c|}{$16\times 20\times 20=6400$}\\
		 \hline
		  & \small Number & \small Percentage & \small Number & \small Percentage\\
		 \hline
		 \textbf{Non-zero} & 1024 & 18.30\% & 788 & 12.31\%\\
		 \hline
		 \textbf{Zeros on the diagonal} & 280 & 5.00\% & 320 & 5.00\%\\
		 \hline
		 \textbf{Missing data} & 3478 & 62.10\% & 4598 & 71.84\%\\
		 \hline
		 \textbf{True zeros} & 818 & 14.60\% & 694 & 10.85\%\\
		 \hline
	\end{tabular}
\end{table}
\subsection{Details on the Datasets Collected}
 The Association of Tennis Professionals (ATP) is the main governing body for male tennis players. The official  ATP website contains records of all matches played on tour. The tournaments of the ATP tour belong to different categories; these include the four Grand Slams, the ATP Masters 1000, etc. The points obtained by the players that ultimately determine their ATP rankings and qualification for entry and seeding in following tournaments depend on the categories of tournaments that they participate or win in. 
% Different categories denote different ranking points players could earn by participating or winning, contributing to the ATP rankings that determine the qualification for entry and seeding in the following tournaments. 
 We selected the most important $M=14$ tournaments for men's dataset, i.e., tournaments that yield the most ranking points which include the four Grand Slams, ATP World Tour Finals and nine ATP Masters 1000, listed in the first column of Table~\ref{tab:temp}. After determining the tournaments, we selected $N=20$ players. We wish to have as many matches as possible between each pair of players, so that the matrices $\{b_{ij}^{(m)}\}, m\in\{1,\ldots,M\}$ would not be too sparse and the algorithm would thus have more data to learn from. We chose players who both have the highest amount of participation in the $M=14$ tournaments from $2008$ to $2017$ and also played the most  number of matches played  in the same period. These players are  listed in the first column of Table~\ref{tab:continue}. 
 
 For each tournament $m$, we collected an $N\times N$ matrix $\{b_{ij}^{(m)}\}$, where $b_{ij}^{(m)}$ denotes the number of times player $i$ beat  player $j$ in tournament $m$. A submatrix consisting of the statistics of matches played between Nadal,  Djokovic,  Federer, and Murray at the French Open is shown in Table~\ref{tab:partial}. We see that over the $10$ years, Nadal beat Djokovic three times and Djokovic beat Nadal once at the French Open. 

 The governing body for women's tennis is the Women's Tennis Association (WTA) instead of the ATP. As such, we collected data from WTA website. The selection of tournaments and players is similar to that for the  men. The tournaments selected include the four Grand Slams, WTA Finals, four WTA Premier Mandatory tournaments, and five Premier 5 tournaments. However, for the first   ``Premier 5'' tournament of the season, the event is either held in Dubai or Doha, and the last tournament was held in Tokyo between 2009 and 2013; this has since been replaced by Wuhan. We decide to treat these two events as four distinct tournaments held   in Dubai, Doha, Tokyo and Wuhan. Hence, the  number of tournaments chosen for the women is $M=16$.

After collecting the data, we checked the sparsity level of the dataset $\mathcal{D}=\{b_{ij}^{(m)}\}$. The zeros in $\mathcal{D}$ can be categorized into three different classes. 
%There are three types of zeros in our datasets where two types of zeros is due to convention and one type of zeros are ``true zeros''.
\begin{enumerate}\itemsep0em
	\item (Zeros on the diagonal) By convention, $b_{ii}^{(m)} = 0$ for all $(i,m)$;
	\item (Missing data) By convention, if player $i$ and $j$ have never played with each other in tournament $m$, then $b_{ij}^{(m)} = b_{ij}^{(m)} = 0$;
	\item (True zeros) If player $i$ has played with player $j$ in tournament $m$ but lost every such match, then  $b_{ij}^{(m)} =0$ and $b_{ji}^{(m)}>0$.
\end{enumerate}
The distributions over the three types of zeros and non-zero entries for male and female players are presented in Table~\ref{tab:sparsity}. We see that there is more missing data in the women's dataset. This is because there has been a small set of dominant male players (e.g., Nadal, Djokovic, Federer) over the past $10$ years but the same is not true for women players. 
For the women, this means that the matches in the past ten years are played by a more diverse set of players, resulting in the number of matches between the top $N=20$ players being smaller compared to the top $N=20$ men, even  though we have selected the same  number  of top players.

\subsection{Running of the Algorithm}

\newcommand*{\MinNumber}{0.0}%
\newcommand*{\RowMaxNumber}{1.0}%
\newcommand*{\ColOneMaxNumber}{1.34E-01}%
\newcommand*{\ColTwoMaxNumber}{1.50E-01}%
\newcommand{\ApplyGradient}[1]{%
        \pgfmathsetmacro{\PercentColor}{70.0*(#1)/\RowMaxNumber} %
        \hspace{-0.33em}\colorbox{red!\PercentColor!white}{#1}
}
\newcommand{\ColOneApplyGradient}[1]{%
        \pgfmathsetmacro{\PercentColor}{70.0*(#1)/\ColOneMaxNumber} %
        \hspace{-0.33em}\colorbox{orange!\PercentColor!white}{#1}
}
\newcommand{\ColTwoApplyGradient}[1]{%
        \pgfmathsetmacro{\PercentColor}{70.0*(#1)/\ColTwoMaxNumber} %
        \hspace{-0.33em}\colorbox{orange!\PercentColor!white}{#1}
}
\newcolumntype{R}{>{\collectcell\ApplyGradient}c<{\endcollectcell}}%
\newcolumntype{C}{>{\collectcell\ColOneApplyGradient}c<{\endcollectcell}}%
\newcolumntype{T}{>{\collectcell\ColTwoApplyGradient}c<{\endcollectcell}}%

\begin{table}[t]
\small
\centering
\caption{Learned dictionary matrix $\mathbf{W}$ for the men's dataset}
\label{tab:temp}
\begin{tabular}{|c|*{1}{R}|*{1}{R}||*{1}{C}|*{1}{T}|}
    \hline
    Tournaments & \multicolumn{2}{c||}{ Row Normalization } & \multicolumn{2}{c|}{Column Normalization}  \\
    \hline
    Australian Open & 5.77E-01	&4.23E-01 & 1.15E-01 & 7.66E-02\\
    Indian Wells Masters & 6.52E-01	&3.48E-01 &1.34E-01	&6.50E-02\\
    Miami Open & 5.27E-01	&4.73E-01 &4.95E-02	&4.02E-02\\
    \cellcolor{gray!40}Monte-Carlo Masters & 1.68E-01	&8.32E-01 &2.24E-02	&1.01E-01\\
    \cellcolor{gray!40}Madrid Open & 3.02E-01	&6.98E-01 &6.43E-02	&1.34E-01\\
    \cellcolor{gray!40}Italian Open & 0.00E-00	&1.00E-00 &1.82E-104	&1.36E-01\\
    \cellcolor{gray!40}French Open & 3.44E-01    &6.56E-01  &8.66E-02	&1.50E-01\\
    Wimbledon & 6.43E-01	&3.57E-01 & 6.73E-02	&3.38E-02\\
    Canadian Open & 1.00E-00	&0.00E-00 &1.28E-01	&1.78E-152\\
    Cincinnati Masters & 5.23E-01	&4.77E-01 &1.13E-01	&9.36E-02\\
    US Open & 5.07E-01	&4.93E-01 &4.62E-02	&4.06E-02\\
    Shanghai Masters & 7.16E-01	&2.84E-01 &1.13E-01	&4.07E-02\\
    Paris Masters & 1.68E-01	&8.32E-01 &1.29E-02	&5.76E-02\\
    ATP World Tour Finals & 5.72E-01	&4.28E-01 &4.59E-02	&3.11E-02\\
    \hline
\end{tabular}
\end{table}

\begin{figure}[t]
	\subfloat[$K=2$ \label{fig:sub1}]{
		\includegraphics[width=0.475\linewidth]{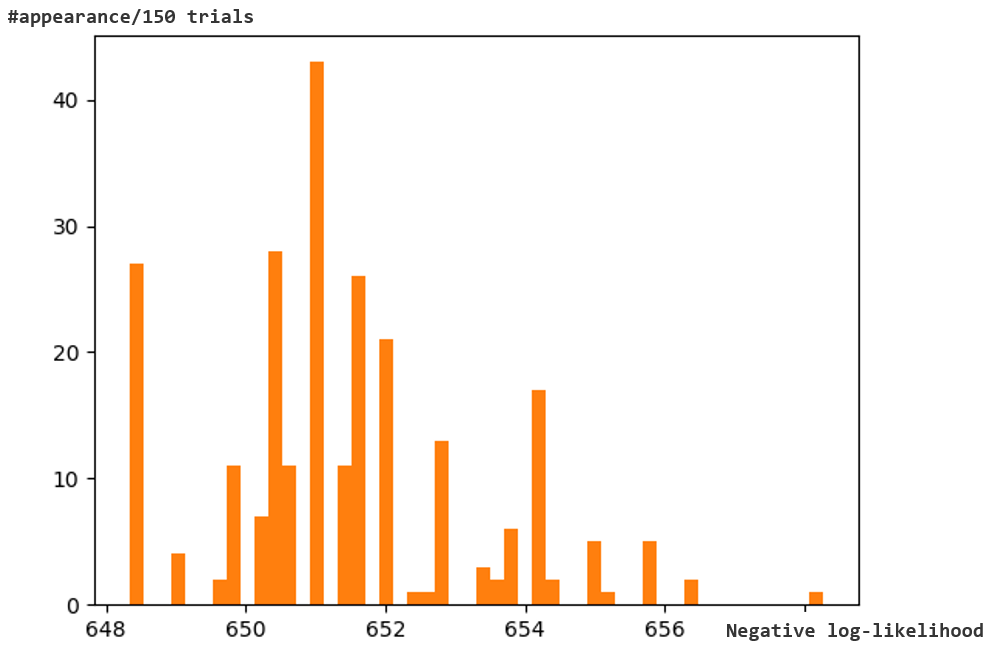}
	}
	\hfill
	\subfloat[$K=3$ \label{fig:sub2}]{
		\includegraphics[width=0.475\linewidth]{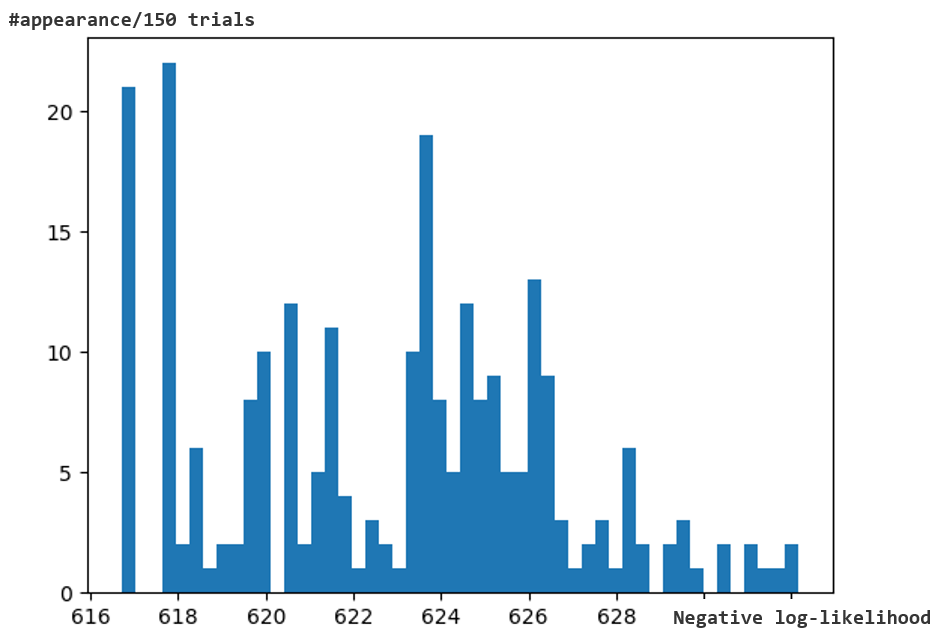}	
	}
	\caption{Histogram of negative log-likelihood in the $150$ trials}
	\label{fig:sub}
\end{figure}

%\begin{wrapfigure}{r}{0.5\textwidth}
%  \begin{center}
%	\includegraphics[scale=0.5]{p3.png}
%  \end{center}
%  \caption{	Plot of the evolution of negative log-likelihoods in one trial} \label{fig:evol}
%\end{wrapfigure}
The number of latent variables is expected to be small and we set $K$  to be~$2$ or~$3$. We only present results for $K=2$  in the main paper; the results corresponding to Tables~\ref{tab:temp} to~\ref{tab:2} for $K=3$ are displayed in Tables S-1 to S-4  in the supplementary material~\cite{xia2019}. We also set $\epsilon=10^{-300}$ which is close to the  smallest positive value in the Python environment. The algorithm terminates when the difference of every element of $\mathbf{W}$ and $\mathbf{H}$ between in the successive  iterations is less than $\tau=10^{-6}$. We checked that the   $\epsilon$-modified algorithm in Sec.~\ref{sec:num} results in non-decreasing likelihoods. See Fig. S-1 in the supplementary material~\cite{xia2019}.
Since~\eqref{eqn:new_ll_func} is non-convex, the MM algorithm can be trapped in local minima. Hence, we considered $150$ different random initializations  for $\mathbf{W}^{(0)}$ and $\mathbf{H}^{(0)}$  and analyzed the result that gave the maximum likelihood  among the $150$ trials. Histograms of the negative log-likelihoods are shown in Figs.~\ref{fig:sub}(a) and \ref{fig:sub}(b) for $K=2$ and $K=3$ respectively. We   observe that the optimal value of the log-likelihood for $K=3$ is  higher than that of $K=2$ since the former model is richer.  We also observe that  the $\mathbf{W}$'s and $\mathbf{H}$'s produced over the $150$ runs are roughly the same up to permutation of rows and columns, i.e., our solution is {\em stable} and  {\em robust} (cf.\ Theorem~\ref{thm:conv} and Sec.~\ref{sec:comp}).

%Following is a histogram showing the negative log-likelihood of 150 trials, where Figure.~\ref{fig:sub1} on the left is when $K=2$ and Figure.~\ref{fig:sub2} on the right is when $K=3$. We can observe that the optimal value of negative log-likelihood of $K=3$ is significantly lower than that of $K=2$, since with more variables the model improves.

\subsection{Results for Men Players}\label{sec:men}

\newcommand*{\PlayerMaxNumber}{1.56E-01}%
\newcommand{\PlayerApplyGradient}[1]{%
        \pgfmathsetmacro{\PercentColor}{70.0*(#1)/\PlayerMaxNumber} %
        \hspace{-0.33em}\colorbox{blue!\PercentColor!white}{#1}
}
\newcolumntype{V}{>{\collectcell\PlayerApplyGradient}c<{\endcollectcell}}%

\begin{table}[t]
\small
\centering
\caption{Learned transpose $\mathbf{H}^T$ of the coefficient matrix for the men's dataset} \label{tab:continue}
\begin{tabular}{|c|*{2}{V|}c|}
    \hline
    Players & \multicolumn{2}{c|}{matrix $\mathbf{H}^{T}$} & Total Matches \\
    \hline
    Novak Djokovic&	1.20E-01&	9.98E-02&  283\\
    Rafael Nadal&	2.48E-02&	1.55E-01&  241 \\
    Roger Federer&	1.15E-01&	2.34E-02&  229\\
    Andy Murray&	7.57E-02&	8.43E-03&  209\\
    Tomas Berdych&	0.00E-00&	3.02E-02&  154\\
    David Ferrer&	6.26E-40&	3.27E-02&  147\\
    Stan Wawrinka&	2.93E-55&	4.08E-02&  141\\
    Jo-Wilfried Tsonga&	3.36E-02&	2.71E-03& 121\\
    Richard Gasquet&	5.49E-03&	1.41E-02& 102\\
    Juan Martin del Potro&	2.90E-02&	1.43E-02& 101\\
    Marin Cilic&  2.12E-02&	0.00E-00& 100\\
    Fernando Verdasco&	1.36E-02&	8.79E-03& 96\\
    Kei Nishikori&	7.07E-03&	2.54E-02& 94\\
    Gilles Simon&	1.32E-02&	4.59E-03&  83\\
    Milos Raonic&	1.45E-02&	7.25E-03& 78\\
    Philipp Kohlschreiber&	2.18E-06&	5.35E-03& 76\\
    John Isner&	2.70E-03&	1.43E-02& 78\\
    Feliciano Lopez&	1.43E-02&	3.31E-03& 75\\
    Gael Monfils&	3.86E-21&	1.33E-02& 70\\
    Nicolas Almagro&	6.48E-03&	6.33E-06& 60\\ 
    \hline
\end{tabular}
\end{table}

The learned dictionary matrix $\mathbf{W}$ is shown in Table~\ref{tab:temp}. 
 In the ``Tournaments'' column, those tournaments whose surface types are known to be clay are highlighted in gray. For ease of visualization, higher values are shaded darker.  If the rows of $\mathbf{W}$ are normalized, we observe that for clay tournaments, the value in the second column is always larger than that in the first, and vice versa. The only exception is the Paris Masters.\footnote{This may be attributed to its position in the seasonal calendar. The Paris Masters is the last tournament before ATP World Tour Finals. Top players  often choose to skip this tournament to prepare for ATP World Tour Finals which is more prestigious. This has led to some surprising results, e.g., David Ferrer, a strong clay player, won the Paris Masters in 2012 (even though the Paris Masters is a hard court indoor tournament).} Since the row sums are equal to $1$, we can interpret the values in the first and second columns of a fixed row as the probabilities that a particular tournament is being played on non-clay or clay surface respectively. If the columns of $\mathbf{W} $ are normalized, it is observed that the  tournaments with highest value of the second column are exactly the four tournaments played on clay. %Since each column of $\mathbf{W}$ sums to one, we can view the second column as probabilities over clay. 
 From $\mathbf{W}$, we learn that surface type---in particular, whether or not a tournament is played on clay---is a germane latent variable that influences the performances of men players. % As a result, we believe that $\mathbf{W}$ has successfully factored out the latent variable - surfaces type.

\newcommand*{\AllMaxNumber}{2.95E-02}%
\newcommand{\AllPlayerApplyGradient}[1]{%
        \pgfmathsetmacro{\PercentColor}{70.0*(#1)/\AllMaxNumber} %
        \hspace{-0.33em}\colorbox{red!\PercentColor!white}{#1}
}
\newcolumntype{Q}{>{\collectcell\AllPlayerApplyGradient}c<{\endcollectcell}}%

\begin{sidewaystable}

\smallskip
\caption{Learned $\mathbf{\Lambda} = \mathbf{W}\mathbf{H}$ matrix for first $10$ men players} \label{tab:1}
\begin{tabular}{|c||*{10}{Q|}}
	\hline
Tournament     &	\multicolumn{1}{p{1.5cm}|}{\tiny Novak Djokovic} 	&	\multicolumn{1}{p{1.5cm}|}{\tiny Rafael Nadal}	&	\multicolumn{1}{p{1.5cm}|}{\tiny Roger Federer}	&	\multicolumn{1}{p{1.5cm}|}{\tiny Andy Murray}	&	\multicolumn{1}{p{1.5cm}|}{\tiny Tomas Berdych}	&	\multicolumn{1}{p{1.5cm}|}{\tiny David Ferrer}	&	\multicolumn{1}{p{1.5cm}|}{\tiny Stan Wawrinka}	&	\multicolumn{1}{p{1.5cm}|}{\tiny Jo-Wilfried Tsonga}	&	\multicolumn{1}{p{1.5cm}|}{\tiny Richard Gasquet}	&	\multicolumn{1}{p{1.5cm}|}{\tiny Juan Martin del Potro}\\
    \hline
    Australian Open	&	2.16E-02	&	1.54E-02	&	1.47E-02	&	9.13E-03	&	2.47E-03	&	2.67E-03	&	3.34E-03	&	3.97E-03	&	1.77E-03	&	4.41E-03	\\
    Indian Wells Masters	&	2.29E-02	&	1.42E-02	&	1.68E-02	&	1.06E-02	&	2.13E-03	&	2.30E-03	&	2.88E-03	&	4.63E-03	&	1.72E-03	&	4.84E-03	\\
    Miami Open	&	2.95E-02	&	2.30E-02	&	1.90E-02	&	1.17E-02	&	3.80E-03	&	4.12E-03	&	5.15E-03	&	5.07E-03	&	2.55E-03	&	5.89E-03	\\
    \cellcolor{gray!40}Monte-Carlo Masters	&	1.19E-02	&	1.53E-02	&	4.46E-03	&	2.27E-03	&	2.90E-03	&	3.14E-03	&	3.92E-03	&	9.12E-04	&	1.46E-03	&	1.94E-03	\\
    \cellcolor{gray!40}Madrid Open	&	1.38E-02	&	1.51E-02	&	6.63E-03	&	3.75E-03	&	2.75E-03	&	2.97E-03	&	3.72E-03	&	1.57E-03	&	1.50E-03	&	2.45E-03	\\
    \cellcolor{gray!40}Italian Open	&	1.19E-02	&	1.84E-02	&	2.78E-03	&	1.00E-03	&	3.59E-03	&	3.89E-03	&	4.87E-03	&	3.23E-04	&	1.68E-03	&	1.71E-03	\\
    \cellcolor{gray!40}French Open	&	1.39E-02	&	1.43E-02	&	7.12E-03	&	4.11E-03	&	2.57E-03	&	2.79E-03	&	3.48E-03	&	1.74E-03	&	1.45E-03	&	2.52E-03	\\
    Wimbledon	&	2.63E-02	&	1.66E-02	&	1.91E-02	&	1.20E-02	&	2.50E-03	&	2.71E-03	&	3.39E-03	&	5.27E-03	&	2.00E-03	&	5.54E-03	\\
    Canadian Open	&	1.16E-02	&	2.40E-03	&	1.11E-02	&	7.32E-03	&	0.00E+00	&	1.26E-39	&	2.42E-51	&	3.25E-03	&	5.31E-04	&	2.81E-03	\\
    Cincinnati Masters	&	1.82E-02	&	1.43E-02	&	1.17E-02	&	7.17E-03	&	2.36E-03	&	2.56E-03	&	3.20E-03	&	3.10E-03	&	1.58E-03	&	3.62E-03	\\
    US Open	&	1.17E-02	&	9.42E-03	&	7.38E-03	&	4.51E-03	&	1.58E-03	&	1.71E-03	&	2.13E-03	&	1.95E-03	&	1.03E-03	&	2.31E-03	\\
    Shanghai Masters	&	8.12E-03	&	4.38E-03	&	6.29E-03	&	4.01E-03	&	6.09E-04	&	6.59E-04	&	8.24E-04	&	1.76E-03	&	5.64E-04	&	1.76E-03	\\
    Paris Masters	&	7.29E-03	&	9.37E-03	&	2.73E-03	&	1.39E-03	&	1.77E-03	&	1.92E-03	&	2.40E-03	&	5.58E-04	&	8.94E-04	&	1.19E-03	\\
    ATP World Tour Finals	&	1.13E-02	&	8.13E-03	&	7.63E-03	&	4.74E-03	&	1.31E-03	&	1.41E-03	&	1.77E-03	&	2.06E-03	&	9.29E-04	&	2.30E-03	\\
	\hline
\end{tabular}

\bigskip\bigskip  % provide some separation between the two tables

\caption{Learned $\mathbf{\Lambda} = \mathbf{W}\mathbf{H}$ matrix for last $10$ men players} \label{tab:2}
\begin{tabular}{|c||*{10}{Q|}}
	\hline
Tournament    &	\multicolumn{1}{p{1.5cm}|}{\tiny Marin Cilic}	&	\multicolumn{1}{p{1.5cm}|}{\tiny Fernando Verdasco}	&	\multicolumn{1}{p{1.5cm}|}{\tiny Gilles Simon}&	\multicolumn{1}{p{1.5cm}|}{\tiny Milos Raonic}	&	\multicolumn{1}{p{1.5cm}|}{\tiny John Isner}	&	\multicolumn{1}{p{1.5cm}|}{\tiny Philipp Kohlschreiber}	&	\multicolumn{1}{p{1.5cm}|}{\tiny John Isner}	&	\multicolumn{1}{p{1.5cm}|}{\tiny Feliciano Lopez}	&	\multicolumn{1}{p{1.5cm}|}{\tiny Gael Monfils}	&	\multicolumn{1}{p{1.5cm}|}{\tiny Nicolas Almagro}\\
    \hline
	Australian Open	&	2.36E-03	&	2.24E-03	&	2.87E-03	&	1.84E-03	&	2.21E-03	&	4.38E-04	&	1.47E-03	&	1.86E-03	&	1.09E-03	&	7.23E-04	\\   
    Indian Wells Masters	&	2.79E-03	&	2.42E-03	&	2.72E-03	&	2.06E-03	&	2.42E-03	&	3.77E-04	&	1.37E-03	&	2.12E-03	&	9.39E-04	&	8.56E-04	\\
    Miami Open	&	2.98E-03	&	3.02E-03	&	4.20E-03	&	2.43E-03	&	2.95E-03	&	6.75E-04	&	2.18E-03	&	2.43E-03	&	1.68E-03	&	9.12E-04	\\
    \cellcolor{gray!40}Monte-Carlo Masters	&	4.10E-04	&	1.11E-03	&	2.58E-03	&	6.96E-04	&	9.77E-04	&	5.14E-04	&	1.43E-03	&	5.95E-04	&	1.28E-03	&	1.26E-04	\\
    \cellcolor{gray!40}Madrid Open	&	8.34E-04	&	1.34E-03	&	2.59E-03	&	9.37E-04	&	1.23E-03	&	4.87E-04	&	1.41E-03	&	8.64E-04	&	1.21E-03	&	2.56E-04	\\
    \cellcolor{gray!40}Italian Open	&	0.00E+00	&	1.05E-03	&	3.03E-03	&	5.47E-04	&	8.63E-04	&	6.38E-04	&	1.71E-03	&	3.95E-04	&	1.59E-03	&	7.68E-07	\\
    \cellcolor{gray!40}French Open	&	9.48E-04	&	1.36E-03	&	2.49E-03	&	9.82E-04	&	1.27E-03	&	4.57E-04	&	1.34E-03	&	9.22E-04	&	1.14E-03	&	2.91E-04	\\
	Wimbledon	&	3.17E-03	&	2.77E-03	&	3.17E-03	&	2.36E-03	&	2.77E-03	&	4.45E-04	&	1.59E-03	&	2.42E-03	&	1.11E-03	&	9.72E-04	\\
	Canadian Open	&	2.05E-03	&	1.32E-03	&	6.84E-04	&	1.27E-03	&	1.40E-03	&	2.26E-07	&	2.62E-04	&	1.38E-03	&	2.46E-19	&	6.27E-04	\\
	Cincinnati Masters	&	1.82E-03	&	1.86E-03	&	2.60E-03	&	1.49E-03	&	1.81E-03	&	4.20E-04	&	1.35E-03	&	1.49E-03	&	1.04E-03	&	5.58E-04	\\
    US Open	&	1.14E-03	&	1.19E-03	&	1.71E-03	&	9.49E-04	&	1.16E-03	&	2.80E-04	&	8.94E-04	&	9.42E-04	&	6.97E-04	&	3.49E-04	\\
    Shanghai Masters	&	1.08E-03	&	8.69E-04	&	8.72E-04	&	7.62E-04	&	8.82E-04	&	1.08E-04	&	4.26E-04	&	7.92E-04	&	2.69E-04	&	3.29E-04	\\
    Paris Masters	&	2.51E-04	&	6.78E-04	&	1.58E-03	&	4.26E-04	&	5.97E-04	&	3.14E-04	&	8.72E-04	&	3.64E-04	&	7.82E-04	&	7.73E-05	\\
    ATP World Tour Finals	&	1.22E-03	&	1.17E-03	&	1.51E-03	&	9.61E-04	&	1.15E-03	&	2.32E-04	&	7.76E-04	&	9.70E-04	&	5.77E-04	&	3.75E-04	\\
	\hline
    \end{tabular}

\end{sidewaystable}

Table~\ref{tab:continue} displays the transpose of $\mathbf{H}$ whose   elements sum to one. Thus, if the column $k\in\{1,2\}$ represents the surface type, we can treat $h_{ki}$ as the skill of  player $i$  conditioned on him playing on surface type $k$. We may regard the first and second columns of $\mathbf{H}^T$ as the skill levels  of players on  non-clay and clay respectively. We   observe that   Nadal, nicknamed the ``King of Clay'', is the best player on clay   among the $N=20$ players, and as an individual, he  is also much more skilful on clay compared to non-clay.    Djokovic, the first  man in the ``Open era'' to hold all four Grand Slams on three different surfaces (hard court, clay and grass) at the same time (between Wimbledon 2015 to the French Open 2016, also known as the Nole Slam), is more of a balanced top player as his skill levels  are high in both columns of $\mathbf{H}^T$.
% but slightly skewed to the non-clay one. Since he is known to be the best player on hard surfaces, this result matches with reality.  
 Federer won the most titles on tournaments played on grass  and, as expected, his skill level in the first column is indeed much higher than the second. As for   Murray, the $\mathbf{H}^T$ matrix also reflects his weakness on clay.   Wawrinka, a player who is known to favor clay has skill level in the second column being much higher than that in the first. The last column of Table~\ref{tab:continue} lists the total number of matches that each player participated in (within our dataset). We verified  that the  skill levels in $\mathbf{H}^T$ for each player are not strongly correlated to  how many matches are being considered in the dataset. Although   Berdych has data of more matches   compared to   Ferrer, his scores are not higher than that of   Ferrer. Thus our algorithm and conclusions are not skewed towards the availability of data.

The learned skill matrix $\mathbf{\Lambda} = \mathbf{W}\mathbf{H}$ with column normalization of $\mathbf{W}$  is presented in Tables~\ref{tab:1} and~\ref{tab:2}. As mentioned in Sec.~\ref{PD}, $[\mathbf{\Lambda}]_{mi}$ denotes the skill level of player $i$ in tournament $m$. We   observe that   Nadal's skill levels are higher than   Djokovic's only for the French Open, Madrid Open, Monte-Carlo Masters, Paris Masters and Italian Open, which are tournaments played on clay  except for the Paris Masters. As for Federer, his skill level is highest for Wimbledon, which happens to be the only tournament on grass; here, it is known that  he is the player with the best record  in the ``Open era''. Furthermore, if we consider Wawrinka,  the five tournaments in which his skill levels are the highest include the four clay tournaments. These   observations again show that our model has learned interesting latent variables from $\mathbf{W}$. It has also learned players' skills on different  types of  surfaces  and tournaments from $\mathbf{H}$ and $\mathbf{\Lambda}$ respectively.

\subsection{Results for Women Players}
\newcommand*{\WRowMaxNumber}{1.0}%
\newcommand*{\WColOneMaxNumber}{1.41E-01}%
\newcommand*{\WColTwoMaxNumber}{1.57E-01}%
\newcommand{\WApplyGradient}[1]{%
        \pgfmathsetmacro{\PercentColor}{70.0*(#1)/\WRowMaxNumber} %
        \hspace{-0.33em}\colorbox{red!\PercentColor!white}{#1}
}
\newcommand{\WColOneApplyGradient}[1]{%
        \pgfmathsetmacro{\PercentColor}{70.0*(#1)/\WColOneMaxNumber} %
        \hspace{-0.33em}\colorbox{orange!\PercentColor!white}{#1}
}
\newcommand{\WColTwoApplyGradient}[1]{%
        \pgfmathsetmacro{\PercentColor}{70.0*(#1)/\WColTwoMaxNumber} %
        \hspace{-0.33em}\colorbox{orange!\PercentColor!white}{#1}
}
\newcolumntype{W}{>{\collectcell\WApplyGradient}c<{\endcollectcell}}%
\newcolumntype{X}{>{\collectcell\WColOneApplyGradient}c<{\endcollectcell}}%
\newcolumntype{U}{>{\collectcell\WColTwoApplyGradient}c<{\endcollectcell}}%

\begin{table}[t]
\small
\centering
\caption{Learned dictionary matrix $\mathbf{W}$ for the women's dataset}
\label{tab:resultwwomen}
\begin{tabular}{|c|*{1}{W}|*{1}{W}||*{1}{X}|*{1}{U}|}
    \hline
    Tournaments & \multicolumn{2}{c||}{ {Row Normalization}} & \multicolumn{2}{c|}{{Column Normalization}} \\
    \hline
    Australian Open & 1.00E-00	&3.74E-26& 1.28E-01	&3.58E-23\\
    Qatar Open & 6.05E-01	&3.95E-01 &1.05E-01	&4.94E-02\\
    Dubai Tennis Championships & 1.00E-00&	1.42E-43 &9.47E-02	&3.96E-39\\
    Indian Wells Open &5.64E-01	&4.36E-01& 8.12E-02	 &4.51E-02 \\
    Miami Open & 5.86E-01	&4.14E-01 &7.47E-02	&3.79E-02\\
    \cellcolor{gray!40}Madrid Open  & 5.02E-01	&4.98E-01 &6.02E-02	&4.29E-02\\
    \cellcolor{gray!40}Italian Open & 3.61E-01&	6.39E-01 &5.22E-02	&6.63E-02\\
    \cellcolor{gray!40}French Open & 1.84E-01	&8.16E-01 &2.85E-02	&9.04E-02\\
    Wimbledon & 1.86E-01	&8.14E-01 &3.93E-02	&1.24E-01\\
    Canadian Open & 4.59E-01	&5.41E-01 &5.81E-02	&4.92E-02\\
    Cincinnati Open & 9.70E-132	&1.00E-00 &5.20E-123	&1.36E-01\\
    US Open & 6.12E-01	&3.88E-01 &8.04E-02	&3.66E-02\\
    Pan Pacific Open & 1.72E-43	&1.00E-00 &7.82E-33	&1.57E-01\\
    Wuhan Open & 1.00E-00	&6.87E-67&1.41E-01	&1.60E-61\\
    China Open & 2.26E-01	&7.74E-01 &4.67E-02	&1.15E-01\\
    WTA Finals & 1.17E-01&	8.83E-01 &9.30E-03	&5.03E-02\\  
    \hline
\end{tabular}

\end{table}

\newcommand*{\WPlayerMaxNumber}{1.44E-01}%
\newcommand{\WPlayerApplyGradient}[1]{%
        \pgfmathsetmacro{\PercentColor}{70.0*(#1)/\WPlayerMaxNumber} %
        \hspace{-0.33em}\colorbox{blue!\PercentColor!white}{#1}
}
\newcolumntype{Y}{>{\collectcell\WPlayerApplyGradient}c<{\endcollectcell}}%

\begin{table}[t]
\small
\centering
\caption{Learned  transpose $\mathbf{H}^T$ of coefficient matrix for the women's dataset}
\label{tab:resulthwomen}
\begin{tabular}{|c|*{2}{Y|}c|}
    \hline
    Players & \multicolumn{2}{c|}{matrix $\mathbf{H}^{T}$} & Total Matches \\
    \hline
Serena Williams&	5.93E-02&	1.44E-01&	130\\
Agnieszka Radwanska&	2.39E-02&	2.15E-02&	126\\
Victoria Azarenka&	7.04E-02&	1.47E-02&	121\\
Caroline Wozniacki&	3.03E-02&	2.43E-02&	115\\
Maria Sharapova&	8.38E-03	&8.05E-02&	112\\
Simona Halep&	1.50E-02	&3.12E-02	&107\\
Petra Kvitova&	2.39E-02	&3.42E-02&	99\\
Angelique Kerber&	6.81E-03&	3.02E-02&	96\\
Samantha Stosur&	4.15E-04&	3.76E-02&	95\\
Ana Ivanovic&	9.55E-03	&2.60E-02&	85\\
Jelena Jankovic&	1.17E-03&	2.14E-02&	79\\
Anastasia Pavlyuchenkova&	6.91E-03	&1.33E-02&	79\\
Carla Suarez Navarro&	3.51E-02&	5.19E-06&	75\\
Dominika Cibulkova&	2.97E-02&	1.04E-02&	74\\
Lucie Safarova&	0.00E+00	&3.16E-02	&69\\
Elina Svitolina&	5.03E-03	&1.99E-02&	59\\
Sara Errani&	7.99E-04	&2.69E-02	&58\\
Karolina Pliskova&	9.92E-03	&2.36E-02&	57\\
Roberta Vinci&	4.14E-02	&0.00E+00	&53\\
Marion Bartoli&	1.45E-02	&1.68E-02	&39\\
    \hline
\end{tabular}
\end{table}
We performed the same experiment for the women players except that we now  consider $M=16$ tournaments.  The factor matrices $\mathbf{W}$ and  $\mathbf{H}$ (in its transpose form) are presented in Tables~\ref{tab:resultwwomen} and~\ref{tab:resulthwomen} respectively. 

It can be seen from   $\mathbf{W}$ that, unlike for the men players,  the surface type is not a pertinent latent variable since there is no correlation between the values in the columns and the surface type. We suspect that the skill levels of top women players are not as heavily influenced by the surface type compared to the men. However, the tournaments in Table \ref{tab:resultwwomen} are ordered in chronological order and we notice that there is a slight correlation between the values in the column and the time of the tournament (first half or second half of the year).   Any latent variable would naturally be less pronounced, due to the sparser dataset for women players (cf.\ Table~\ref{tab:sparsity}).  A somewhat interesting observation is that the values in $\mathbf{W}$ obtained using  the row normalization and the column normalization methods are similar. This indicates that the latent variables, if any, learned by the two methods are the same, which is a reassuring conclusion.

By computing the sums of the skill levels for each female player (i.e., row sums of  $\mathbf{H}^T$), we   see that S.~Williams is the most skilful   among the $20$ players over the past $10$ years. She is followed by  Sharapova and Azarenka.  As a matter of fact, S.~Williams and Azarenka have been year-end number one $4$ times and once, respectively, over the period $2008$ to~$2017$. Even though Sharapova was never at the top at the  end of any season (she was, however, ranked number one several times, most recently in 2012),  she had been consistent over this period such that the model and the longitudinal dataset allow us to  conclude that she is ranked second. In fact, she is known for her unusual longevity being at the top of the women's game. She started her tennis career very young and won her first Grand Slam at the age of $17$. Finally, the model groups S.~Williams, Sharapova, Stosur together, while Azarenka, Navarro, and Vinci are in another group. We believe that there may be some similarities between players who are clustered in the same group. The $\mathbf{\Lambda}$ matrix for women players can be found in Tables S-5 and S-6 in the supplementary material~\cite{xia2019}.

%Result of $\mathbf{W}$ and $\mathbf{H}$ for women's dataset are presented in Table.\ref{tab:resultwwomen} and Table.\ref{tab:resulthwomen}. Unfortunately, the latent variables factored out by the model do not include surface type since there is no pattern to be observed between the values and whether tournaments are with clay surfaces. We suspect that women do not have very significant difference of skills of playing on different surfaces as compared with men, maybe other variables influence women more. A reasonable discovery is that the numbers of $\mathbf{W}$ shown by the row normalization method and the column normalization method are similar, which indicates that the latent variables the two method used are same. By looking at the result of $\mathbf{H}$ for women's result, we can still observe that Serena Williams are the best female player among the 20, followed by the player Maria Sharapova. The model somehow indicates grouping Serena Williams, Maria Sharapova and Samantha Stosur together, while Victoria Azarenka, Carla Suarez Navarro and Roberta Vinci in another group. We believe that there are some similarities between the players who are clustered into the same group.
\newcommand*{\BPlayerMaxNumber}{1.35E-01}%
\newcommand*{\BOPlayerMaxNumber}{2.14E-01}%
\newcommand{\BPlayerApplyGradient}[1]{%
        \pgfmathsetmacro{\PercentColor}{70.0*(#1)/\BPlayerMaxNumber} %
        \hspace{-0.33em}\colorbox{red!\PercentColor!white}{#1}
}
\newcolumntype{D}{>{\collectcell\BPlayerApplyGradient}c<{\endcollectcell}}%
\newcommand{\BOPlayerApplyGradient}[1]{%
        \pgfmathsetmacro{\PercentColor}{70.0*(#1)/\BOPlayerMaxNumber} %
        \hspace{-0.33em}\colorbox{orange!\PercentColor!white}{#1}
}
\newcolumntype{I}{>{\collectcell\BOPlayerApplyGradient}c<{\endcollectcell}}%

\begin{table}[t]
\small
\centering
\caption{Learned $\bm{\lambda}$'s for the BTL ($K=1$) and mixture-BTL ($K=2$) models}
\label{tab:mixturebtl}
\begin{tabular}{|c|*{1}{I}|*{1}{D}|*{1}{D}|}
    \hline
\textbf{Players}  & \multicolumn{1}{c|}{$K=1$} & \multicolumn{2}{c|}{$K=2$} \\
    \hline
Novak Djokovic	&	2.14E-01	&	7.14E-02	&	1.33E-01	\\
Rafael Nadal	&	1.79E-01	&	1.00E-01	&	4.62E-02	\\
Roger Federer	&	1.31E-01	&	1.35E-01	&	1.33E-02	\\
Andy Murray	&	7.79E-02	&	6.82E-02	&	4.36E-03	\\
Tomas Berdych	&	3.09E-02	&	5.26E-02	&	2.85E-04	\\
David Ferrer	&	3.72E-02	&	1.79E-02	&	4.28E-03	\\
Stan Wawrinka	&	4.32E-02	&	2.49E-02	&	4.10E-03	\\
Jo-Wilfried Tsonga	&	2.98E-02	&	3.12E-12	&	1.08E-01	\\
Richard Gasquet	&	2.34E-02	&	1.67E-03	&	2.97E-03	\\
Juan Martin del Potro	&	4.75E-02	&	8.54E-05	&	4.85E-02	\\
Marin Cilic	&	1.86E-02	&	3.37E-05	&	2.35E-03	\\
Fernando Verdasco	&	2.24E-02	&	5.78E-02	&	8.00E-09	\\
Kei Nishikori	&	3.43E-02	&	5.37E-08	&	3.58E-02	\\
Gilles Simon	&	1.90E-02	&	7.65E-05	&	5.16E-03	\\
Milos Raonic	&	2.33E-02	&	2.61E-04	&	6.07E-03	\\
Philipp Kohlschreiber	&	7.12E-03	&	1.78E-25	&	3.55E-03	\\
John Isner	&	1.84E-02	&	2.99E-02	&	1.75E-08	\\
Feliciano Lopez	&	1.89E-02	&	1.35E-02	&	3.10E-04	\\
Gael Monfils	&	1.66E-02	&	5.38E-10	&	6.53E-03	\\
Nicolas Almagro	&	7.24E-03	&	1.27E-15	&	1.33E-03	\\
    \hline
\textbf{Mixture weights} & \multicolumn{1}{c|}{1.00E+00} & \multicolumn{1}{c|}{4.72E-01}	&	\multicolumn{1}{c|}{5.28E-01}  \\
 	\hline 
 	\textbf{Log-likelihoods} & \multicolumn{1}{c|}{-682.13} & \multicolumn{2}{c|}{-657.56}  \\
 	\hline
\end{tabular}

\end{table}

\subsection{Comparison to BTL and mixture-BTL} \label{sec:comp}
 Finally, we compared our approach to the   BTL and mixture-BTL~\cite{Oh2014,NiharSimpleRobust} approaches for the male players. To learn these models, we  aggregated our dataset $\{ b_{ij}^{(m)}\}$  into a single matrix $\{b_{ij} = \sum_m b_{ij}^{ (m) }\}$. For the BTL model, we maximized the likelihood  to find the optimal parameters. For the mixture-BTL model with $K=2$ components, we ran an Expectation-Maximization (EM) algorithm~\cite{Demp} to find approximately-optimal values of the parameters and the mixture weights.  Note that the BTL model corresponds to a mixture-BTL model with $K=1$. 
 
 The learned skill vectors are shown in Table~\ref{tab:mixturebtl}.  Since EM is susceptible to being trapped in local optima and is sensitive to initialization, we ran it $100$ times and reported the solution with likelihood that is close to the  highest one.\footnote{The solution with the highest likelihood is shown in Trial 2 of Table S-7 but it appears that the solution there is degenerate.}  The solution for mixture-BTL is not stable; other solutions with likelihoods that are very close to the maximum one result in significantly different parameter values.  Two other solutions with similar likelihoods are shown in Table S-7 in the supplementary material~\cite{xia2019}. As can be seen, some of the solutions  are far from representative of the true skill levels of the players (e.g., in Trial 2 of Table S-7,  Tsonga has a very high score in the first column and the skills of the other players are all very small in comparison) and they are vastly different from one another. This is in  stark contrast to our BTL-NMF model and algorithm in which Theorem~\ref{thm:conv} states that the limit of $\{ (\mathbf{W}^{(l)}, \mathbf{H}^{(l)})\}_{l=1}^\infty$ is a stationary point of~\eqref{eqn:min}. We numerically verified that the BTL-NMF solution is stable, i.e., different runs yield $(\mathbf{W},\mathbf{H})$ pairs that are approximately equal up to permutation of rows and columns.\footnote{Note, however, that   stationary points are not necessarily equivalent up to permutation or rescaling.}  As seen from Table~\ref{tab:mixturebtl}, for mixture-BTL, neither tournament-specific information nor semantic meanings of latent variables can be gleaned from the   parameter vectors. The results of BTL are reasonable and expected but also lack tournament-specific information.
\section{Conclusion and Future Work}\label{sec:con}
%In this paper, 
We proposed a ranking model combining the BTL model with the NMF framework as in Fig.~\ref{fig:nmf_btl}. We derived MM-based algorithms to maximize the likelihood of the data. To ensure numerical stability, we ``regularized'' the MM algorithm and proved that  desirable properties, such as monotonicity of the objective and convergence of the iterates to stationary points, hold. We drew interesting conclusions based on longitudinal datasets for top male and female players.  A latent variable in the form of the court surface was also uncovered in a principled manner. We compared our approach to the mixture-BTL approach~\cite{Oh2014,NiharSimpleRobust}  and concluded that the former is advantageous in various aspects (e.g., stability of the solution, interpretability of latent variables).

In the future, we plan to run our algorithm on a larger longitudinal dataset consisting of pairwise comparison data from more years (e.g., the past $50$ years) to learn, for example, who is the ``best-of-all-time'' male or female player. In addition, it would be desirable to understand if there is a natural Bayesian interpretation~\cite{tan2013automatic,CaronDoucet} of the $\epsilon$-modified objective function in~\eqref{eqn:new_ll_func}.

\paragraph{Acknowledgements} This work was supported by a Singapore Ministry of Education Tier 2 grant (R-263-000-C83-112), a  Singapore National Research Foundation (NRF) Fellowship (R-263-000-D02-281), and by the European Research Council (ERC FACTORY-CoG-6681839).

%This work was supported in part by a Singapore Ministry of Education Tier 2 grant (R-263-000-C83-112), in part by an NRF Fellowship (R-263-000-D02-281). 
 % and using majorization-minimizat-ion algorithm. The model is tested on tennis dataset for both male and female. The result for male tennis showed the model successfully factored out latent variables that have physical interpretation which is surface type. Numerical problems are carefully and systematically solved together with proper and meaningful normalization constraints. 

\bibliographystyle{unsrt}
\bibliography{isitbib}

\begin{thebibliography}{10}

\bibitem{EloBook}
A.~E. Elo.
\newblock {\em The Rating of Chess Players, Past and Present}.
\newblock Ishi Press International, 2008.

\bibitem{bradleyterry}
R.~Bradley and M.~Terry.
\newblock Rank analysis of incomplete block designs { I}: The method of paired
  comparisons.
\newblock {\em Biometrika}, 35:324--345, 1952.

\bibitem{Luce}
R.~Luce.
\newblock {\em Individual choice behavior: A theoretical analysis}.
\newblock Wiley, 1959.

\bibitem{LS99}
D.~D. Lee and H.~S. Seung.
\newblock {Learning the parts of objects with nonnegative matrix
  factorization}.
\newblock {\em Nature}, 401:788--791, 1999.

\bibitem{CichockiBook}
A.~Cichocki, R.~Zdunek, A.~H. Phan, and S.-I. Amari.
\newblock {\em Nonnegative Matrix and Tensor Factorizations: Applications to
  Exploratory Multi-Way Data Analysis and Blind Source Separation}.
\newblock John Wiley \& Sons, Ltd, 2009.

\bibitem{MardenBook}
J.~I. Marden.
\newblock {\em Analyzing and modeling rank data}.
\newblock CRC Press, 1996.

\bibitem{leeseung2000}
D.~D. Lee and H.~S. Seung.
\newblock Algorithms for nonnegative matrix factorization.
\newblock In {\em Neural Information Processing Systems}, pages 535--541, 2000.

\bibitem{fevotte2009}
C.~F\'evotte, N.~Bertin, and J.-L. Durrieu.
\newblock Nonnegative matrix factorization with the {Itakura-Saito} divergence.
  with application to music analysis.
\newblock {\em Neural Computation}, 21(3):793--830, Mar 2009.

\bibitem{berry2005}
M.~W. Berry and M.~Browne.
\newblock Email surveillance using non-negative matrix factorization.
\newblock {\em Computational and Mathematical Organization Theory},
  11(3):249--264, 2005.

\bibitem{geerts2018}
A.~Geerts, T.~Decroos, and J.~Davis.
\newblock Characterizing soccer players' playing style from match event
  streams.
\newblock In {\em Machine Learning and Data Mining for Sports Analytics
  ECML/PKDD 2018 workshop}, pages 115--126, 2018.

\bibitem{hunterLange2004}
D.~R. Hunter and K.~Lange.
\newblock A tutorial on {MM} algorithms.
\newblock {\em American Statistician}, 58:30--37, 2004.

\bibitem{zhao2017unified}
R.~Zhao and V.~Y.~F. Tan.
\newblock A unified convergence analysis of the multiplicative update algorithm
  for regularized nonnegative matrix factorization.
\newblock {\em IEEE Transactions on Signal Processing}, 66(1):129--138, 2018.

\bibitem{Raz}
M.~Razaviyayn, M.~Hong, and Z.-Q. Luo.
\newblock A unified convergence analysis of block successive minimization
  methods for nonsmooth optimization.
\newblock {\em SIAM Journal on Optimization}, 23(2):1126--1153, 2013.

\bibitem{Oh2014}
S.~Oh and D.~Shah.
\newblock Learning mixed multinomial logit model from ordinal data.
\newblock In {\em Neural Information Processing Systems}, pages 595--603, 2014.

\bibitem{NiharSimpleRobust}
N.~B. Shah and M.~J. Wainwright.
\newblock Simple, robust and optimal ranking from pairwise comparisons.
\newblock {\em Journal of Machine Learning Research}, 18(199):1--38, 2018.

\bibitem{Suh17}
C.~Suh, V.~Y.~F. Tan, and R.~Zhao.
\newblock Adversarial top-{$K$} ranking.
\newblock {\em IEEE Transactions on Information Theory}, 63(4):2201--2225,
  2017.

\bibitem{ding2015}
W.~Ding, P.~Ishwar, and V.~Saligrama.
\newblock A topic modeling approach to ranking.
\newblock In {\em Proceedings of the 18th International Conference on
  Artificial Intelligence and Statistics (AISTATS)}, pages 214--222, 2015.

\bibitem{fevotte2011algorithms}
C.~F{\'e}votte and J.~Idier.
\newblock Algorithms for nonnegative matrix factorization with the
  $\beta$-divergence.
\newblock {\em Neural Computation}, 23(9):2421--2456, 2011.

\bibitem{tan2013automatic}
V.~Y.~F. Tan and C.~F\'evotte.
\newblock Automatic relevance determination in nonnegative matrix factorization
  with the $\beta$-divergence.
\newblock {\em IEEE Transactions on Pattern Analysis and Machine Intelligence},
  35(7):1592--1605, 2013.

\bibitem{hunter2004mm}
D.~R. Hunter.
\newblock {MM} algorithms for generalized {Bradley-Terry} models.
\newblock {\em The Annals of Statistics}, 32(1):384--406, 2004.

\bibitem{xia2019}
R.~Xia, V.~Y.~F. Tan, L.~Filstroff, and C.~F\'evotte.
\newblock Supplementary material for ``{A} ranking model motivated by
  nonnegative matrix factorization with applications to tennis tournaments''.
\newblock \url{https://github.com/XiaRui1996/btl-nmf}, 2019.

\bibitem{Demp}
A.~P. Dempster, N.~M. Laird, and D.~B. Rubin.
\newblock Maximum likelihood from incomplete data via the {EM} algorithm.
\newblock {\em Journal of the Royal Statistical Society B}, 39(1--38), 1977.

\bibitem{CaronDoucet}
F.~Caron and A.~Doucet.
\newblock Efficient {Bayesian} inference for generalized {Bradley-Terry}
  models.
\newblock {\em Journal of Computational and Graphical Statistics},
  21(1):174--196, 2012.

\end{thebibliography}

\end{document}